\documentclass{article}

    \PassOptionsToPackage{numbers, compress}{natbib}

 \usepackage[preprint]{neurips_2026}

\usepackage[utf8]{inputenc} %
\usepackage[T1]{fontenc}    %
\usepackage{hyperref}       %
\usepackage{url}    

\usepackage{booktabs}       %
\usepackage{amsfonts}       %
\usepackage{nicefrac}       %
\usepackage{subcaption}
\usepackage{microtype}      %
\usepackage{xcolor}         %
\usepackage{amsmath}
\usepackage{booktabs}
\usepackage{longtable}
\usepackage{graphicx}    
\usepackage{wrapfig}
\usepackage{cleveref}
\usepackage{url}
\usepackage{authblk}
\usepackage{array}
\usepackage[normalem]{ulem}
\newcommand{\pgraph}[1]{\textbf{#1}\,\,}

\title{Understanding the Surprising Generalization Properties of Tabular Foundation Models}

\author{
\textbf{Nour Shaheen}$^{1,2,3,\dag}$,
\textbf{Junwei Ma}$^{4,\dag}$,
\textbf{Alex Labach}$^{5}$,
\textbf{Frank Hutter}$^{6,7,8}$,
\textbf{Valentin Thomas}$^{9}$,
\textbf{Anthony L. Caterini}$^{5}$\\
$^{1}$Polytechnique Montréal,
$^{2}$Mila -- Quebec AI Institute,
$^{3}$Chandar Research Lab,\\
$^{4}$University of Toronto,
$^{5}$Layer 6 AI,
$^{6}$Prior Labs,
$^{7}$ELLIS Institute Tübingen,\\
$^{8}$University of Freiburg,
$^{9}$Cohere
}

\begin{document}

\maketitle
\begingroup
\renewcommand{\thefootnote}{\dag}
\footnotetext[1]{Equal contribution.}
\endgroup

\begingroup
\renewcommand\thefootnote{\fnsymbol{footnote}}
\endgroup

\vspace{-2em}

\begin{abstract}
\looseness = -1 Tabular Foundation Models (TFMs) increasingly rely on in-context learning, where a model receives labelled examples at inference time and predicts labels for new inputs without updating its weights. Existing TFMs are typically trained on either massive synthetic corpora or very large collections of real datasets. In contrast, we show that surprisingly strong transfer can emerge from self-supervised pre-training on just a \emph{single} real table. In this setting, we also find that tables tend to be either broadly useful or broadly poor regardless of downstream prediction task, and that the strongest predictor of usefulness is the number of features rather than the number of instances. This leads to a task-centric interpretation of tabular pre-training: the number and the quality of %
tasks are essential for the pre-training of TFMs.
We show that the same task-centric perspective can help corpus design at scale: fine-grained column-level pre-processing consistently improves downstream performance, while no improvements are observed when we filter or deduplicate at the dataset level.
Finally, we offer a new perspective for how TFMs generalize: we believe that tabular in-context generalization is largely retrieval-based, and good models are those that learn to identify relevant examples in the provided context and aggregate them well. The mechanics of TFMs have been relatively understudied; our task-centric, retrieval-based perspective offers a new framework to guide future model and corpus design.

\end{abstract}

\vspace{-1em}

\section{Introduction}

\looseness=-1  Tabular data is commonly viewed as highly heterogeneous, leading to a prevailing belief that transfer learning across tabular domains is nearly impossible: e.g., \emph{how could training on handwritten digits possibly transfer to estimating Californian housing prices?} 
However, recent tabular in‑context learning (ICL) methods such as TabPFN~\citep{hollmann2023tabpfn,Hollmann2025}, TabDPT~\citep{ma2025tabdptscalingtabularfoundation}, and TabICL~\citep{qu2025tabicl,qu2026tabiclv2}---termed tabular foundation models (TFMs)---have challenged this assumption. These methods train transformers using either massive amounts of synthetic data (millions of tables generated from structured priors~\citep{hollmann2023tabpfn,Hollmann2025,qu2025tabicl,qu2026tabiclv2}) or by sampling from large collections of real tabular datasets~\citep{ma2025tabdptscalingtabularfoundation,garg2025real}, randomizing context composition and prediction targets at each step to encourage in-context generalization.
All modern tabular ICL models of which we are aware (e.g.,~\citep{hollmann2023tabpfn,Hollmann2025,ma2025tabdptscalingtabularfoundation, qu2025tabicl, qu2026tabiclv2, zhang2025limix, gardner2024large, spinaci2025contexttab, zhang2025mitra}) share a similar task construction procedure during pre-training: a dataset is either selected or created, a column of this dataset is used as the target, and a subset of the remaining columns are used as features.
Inference is then performed on new datasets while keeping the weights frozen: %
labelled examples from the evaluation dataset are given as context and the model predicts the values of unlabelled examples.
The general belief is that, much like in large language models, only a very diverse pre-training set can cover the distribution of unseen tabular tasks and enable out-of-domain generalization~\citep{muller2022transformers,hollmann2023tabpfn}.

\begin{figure}[t]
    \centering
    \includegraphics[width=\textwidth]{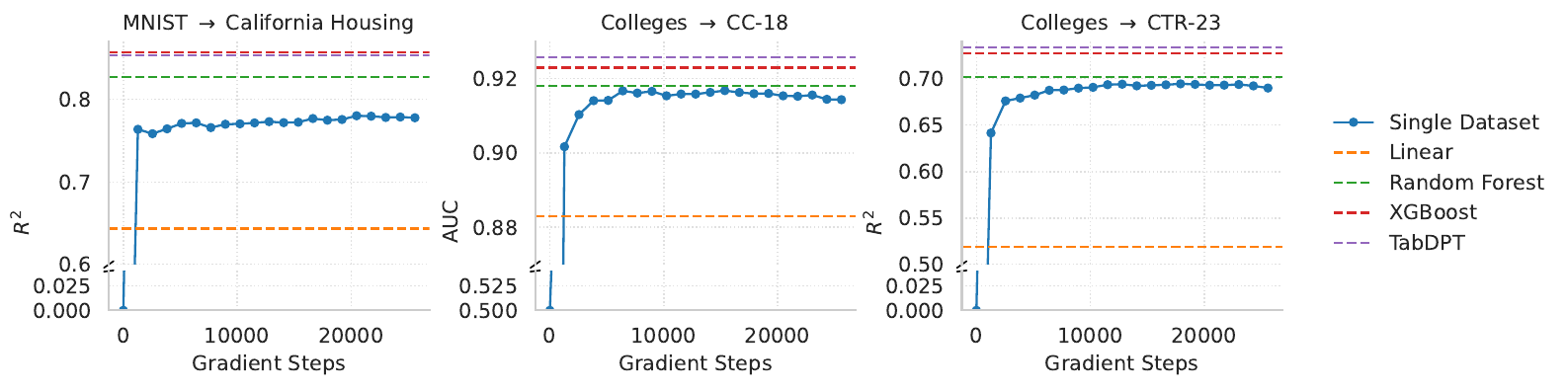}
    \caption{
        Transfer from a single pre‑training dataset. 
        \textbf{Left:} Training only on vectorized \textsc{MNIST} (treated as a table) and evaluating on \textsc{California Housing}.
        \textbf{Middle and Right:} Training on the \textsc{Colleges} dataset and evaluating on the full CC-18 and CTR-23 evaluation suites, respectively.
    }
    \label{fig:transfer_mnist2_california}
\end{figure}

\looseness=-1
Yet we find the opposite, and this surprising finding underpins our work: even with a \emph{single} real-world table---such as vectorized \textsc{MNIST}~\citep{lecun1998gradient}%
---a transformer trained with a na\"ive self-supervised learning (SSL) objective
can still acquire generalization capabilities that transfer robustly \emph{across domains}. 
Figure~\ref{fig:transfer_mnist2_california} illustrates this: a transformer trained from scratch only on the \textsc{MNIST} table yields strong in-context performance on structurally and semantically unrelated datasets like \textsc{California Housing}~\citep{pace1997sparse}. Similarly, training on the \textsc{Colleges} dataset~\citep{OpenML_dataset_42727} and evaluating the resulting model on the OpenML CC-18~\citep{bischl2021cc18} and CTR-23~\citep{fischer2023ctr} benchmark suites (CC-BY license) ends up closely matching the performance of a random forest baseline.

\looseness=-1
In this paper, we first investigate which properties of single real-world datasets are most linked to generalization. We find: (1) Datasets that transfer well tend to do so universally across evaluation sets; (2) Feature count is a strong predictor of generalization, while instance count matters little; and (3) The number of \emph{unique tasks} a model is pre-trained on strongly drives generalization ability.

\looseness=-1
We then apply this insight to pre-training on large data corpora.
The task-centric view still applies with minor tweaks to control for data quality: more tasks generally help, as long as they have \emph{sufficient quality}.
In particular, removing duplicated, excessively correlated, or otherwise uninformative columns from each dataset further improves performance over baselines.
Conversely, dataset-level filtering---even deduplication---is ineffective in our experiments.
Although one might expect datasets resembling the ``bad'' single pre-training datasets to hurt performance, we find the opposite: adding them to ``good'' datasets still tends to improve generalization.

\looseness=-1
Finally, we provide an explanation for what makes generalization possible for TFMs.
We believe the original Bayesian interpretation~\citep{muller2022transformers,hollmann2023tabpfn} is unsatisfying since it requires downstream tasks to be covered by the distribution of the prior, which directly contradicts single-table generalization.
We instead argue that TFMs act as learned retrieval-and-aggregation procedures.
We provide experimental evidence for this by showing (1) a strong link between a model's performance and its ability to retrieve points within its own context, and (2) that architecture modifications enforcing nearest-neighbour-like behaviour have limited impacts on performance.
We also show that the strongest single-dataset models display similar patterns in their attention maps, suggesting (although not proving) the existence of a universal retrieval space enabling generalization.

\section{Generalization from single table training}
\label{sec:single-table}

\looseness=-1
We design experiments to explore the central question of what constitutes a ``good dataset'' such that a tabular ICL model pre-trained on it can generalize.
To isolate the effects of the training datasets themselves, we fix both the model architecture and the pre-training procedure throughout most experiments.
Specifically, we use the shared backbone architecture from TabPFNv1~\citep{hollmann2023tabpfn} and TabDPT~\citep{ma2025tabdptscalingtabularfoundation}, a widely adopted design for tabular ICL. 
We use the \href{https://github.com/layer6ai-labs/TabDPT-training}{pre-training procedure from TabDPT} to train tabular ICL models from scratch using a single real dataset only; as explained above, this procedure is fundamentally similar to all standard TFM training protocols.
To isolate the effect of data, we choose \emph{not} to use retrieval mechanisms (as used in TabDPT) during pre-training in this section, although it is enabled during inference.
We evaluate the generalization performance on established benchmarks in order to easily compare with reported baseline performances directly.

\subsection{Are good pre-training datasets universally good?} \label{sec:universally_good}

\looseness=-1  In~\Cref{fig:transfer_mnist2_california}, we showed the surprising finding that pre-training an ICL tabular model on \textsc{MNIST} can yield strong performance on unrelated datasets.
We ask: \emph{``Is each pre-training dataset only good for some downstream task or does it generalize universally?''}
We pre-train models separately on each of $N_\text{train} = 88$ training datasets from the TabDPT pre-training data, spanning different modalities and sizes, and evaluate the resulting $88$ models on $N_\text{eval} = 107$ diverse datasets: the $72$ CC-18~\citep{bischl2021cc18} classification datasets and $35$ CTR-23~\citep{fischer2023ctr} regression datasets.
No dataset appears in both pre-training and evaluation.
Linear and RandomForest baselines were computed using \texttt{scikit-learn} defaults, except with logistic regression maximum iterations increased to 1000. %
 XGBoost is the tuned baseline from~\citet{mcelfresh2023neural}, and TabDPT is computed from the \href{https://github.com/layer6ai-labs/TabDPT-inference}{public repository} (Version 1.1, Apache-2.0 license).

\begin{figure}[ht]
    \centering
    \begin{subfigure}[b]{0.45\textwidth}
        \centering
        \includegraphics[width=0.99\textwidth]{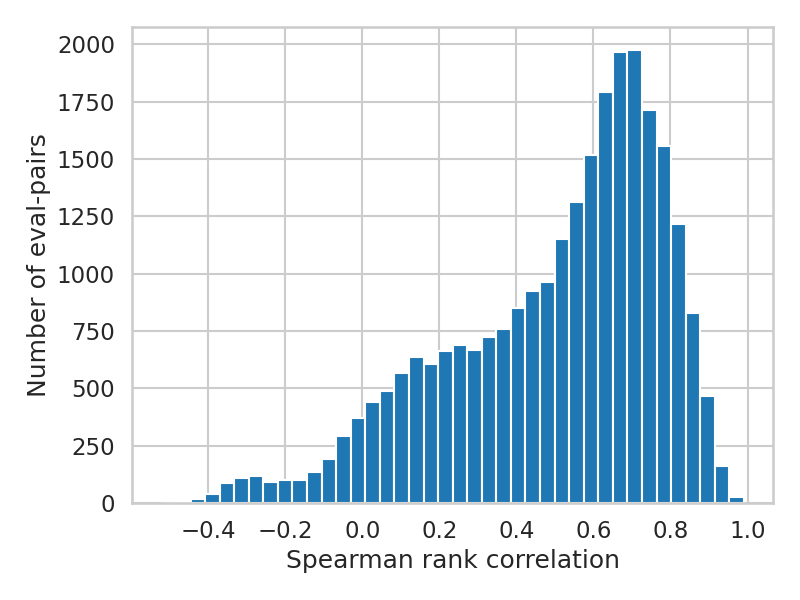}
        \caption{Ranking Correlation: The high values indicate that pre-training datasets which are good on one evaluation dataset tend to be good on others too.}
        \label{fig:stable_ranks}
    \end{subfigure}
    \hfill
    \begin{subfigure}[b]{0.45\textwidth}
        \centering
        \includegraphics[width=0.99\textwidth]{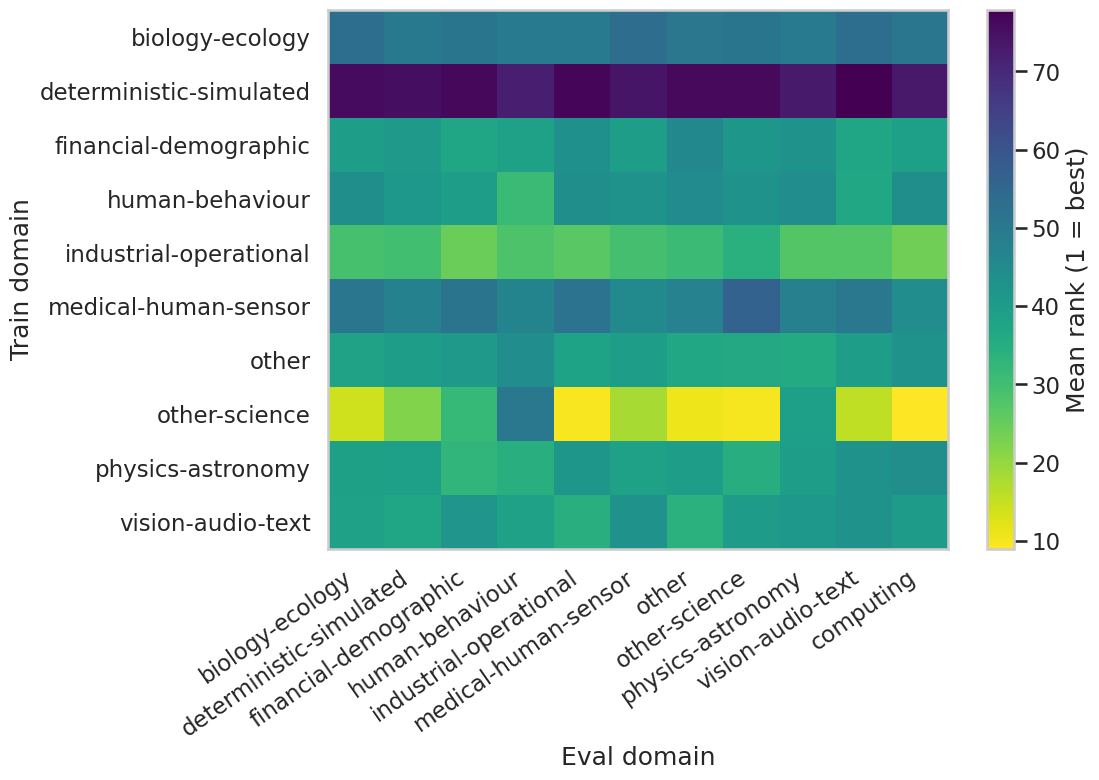}
        \caption{Domain to Domain Transfer: We do not observe stronger transfer when the pre-training and evaluation domain are shared.}
        \label{fig:domain2domain}
    \end{subfigure}
    \caption{\looseness=-1 Universality of dataset quality. \textbf{(a)} We compute the rank of training sets for each evaluation dataset and then plot a histogram of the Spearman correlation between the ranks on all pairs of evaluation datasets. Most evaluation datasets have very correlated ranks. \textbf{(b)} We group the training and evaluation datasets into distinct domains and plot average ranks (lower is better). Training and evaluation pairs from the same domain do not appear to transfer better than ones from different domains.}
    \label{fig:universal}
\end{figure}

\looseness=-1
In~\Cref{fig:stable_ranks}, for each evaluation dataset, we rank the pre-training datasets based on their performance on that task, leading to $N_{\text{eval}}$ rankings.
We compute Spearman correlations over all ranking pairs, with higher values indicating stable relative rankings.
Since the histogram is heavily left-skewed, it indicates that if a dataset is good for one evaluation dataset it tends to be good for others too, \emph{even across vastly different evaluation tasks}.
Next, we investigate whether a training dataset tends to generalize better to evaluation datasets within the same \emph{domain}.
We manually categorize both pre-training and evaluation datasets into domains, record the final performance (either accuracy for classification or correlation for regression) for each single training dataset, and then rank the training datasets from best to worst for each domain. Finally, we aggregate over the domains of the training datasets.
If datasets from, e.g., Domain A are expected to transfer, on average, better to Domain A, we would expect a strong diagonal dominance in this matrix.
However, in~\Cref{fig:domain2domain}, we can see this is not the case: within-domain generalization is not higher. Rather, across the board, some domains are just broadly better for pre-training (e.g., \texttt{other-science} is great, \texttt{deterministic-simulated} is not).

\subsection{What constitutes a good dataset?}

To understand which properties of a training dataset yield successful generalization, we run a meta-analysis on the $N_\text{train}$ pre-training datasets.
For each dataset, we collect descriptive meta-features, including: the number of features, number of instances, number of categorical features, number of numeric features, amount of missing values, and final pre-training losses for both classification and regression heads.
These meta-features are split into meta-train ($80\%$) and meta-test ($20\%$) sets.

\begin{figure}[ht]
    \centering
    \vspace{-1em}
    \begin{subfigure}[t]{0.42\textwidth}
        \centering
    \includegraphics[width=0.99\textwidth]{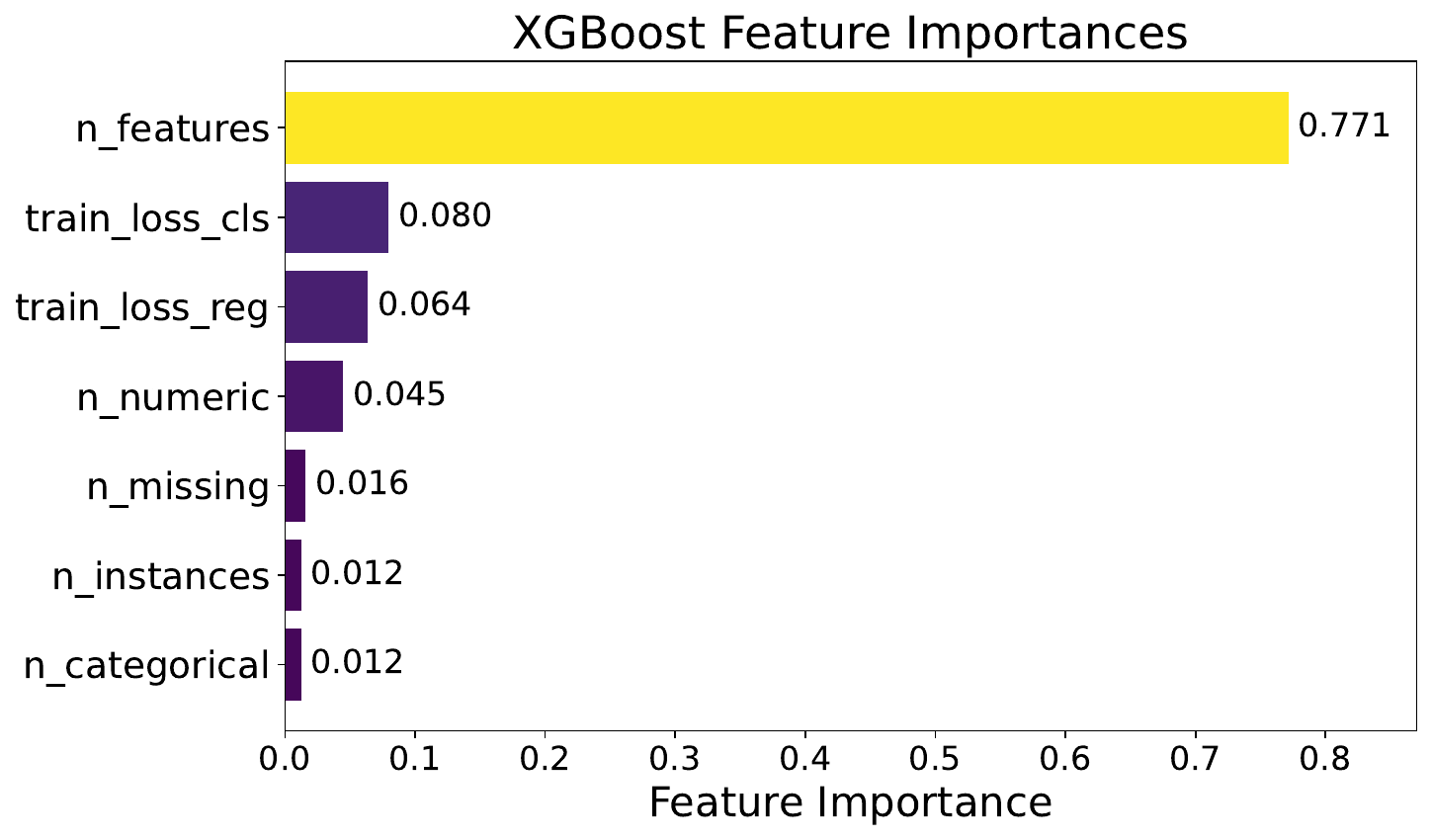}
    \caption{
        Feature importance from an XGBoost model trained to predict average performance across 107 downstream tasks from dataset metadata. 
    }
    \label{fig:xgb_feature_importance}
    \end{subfigure}
    \hfill
    \begin{subfigure}[t]{0.45\textwidth}
        \centering
        \includegraphics[width=0.99\textwidth]{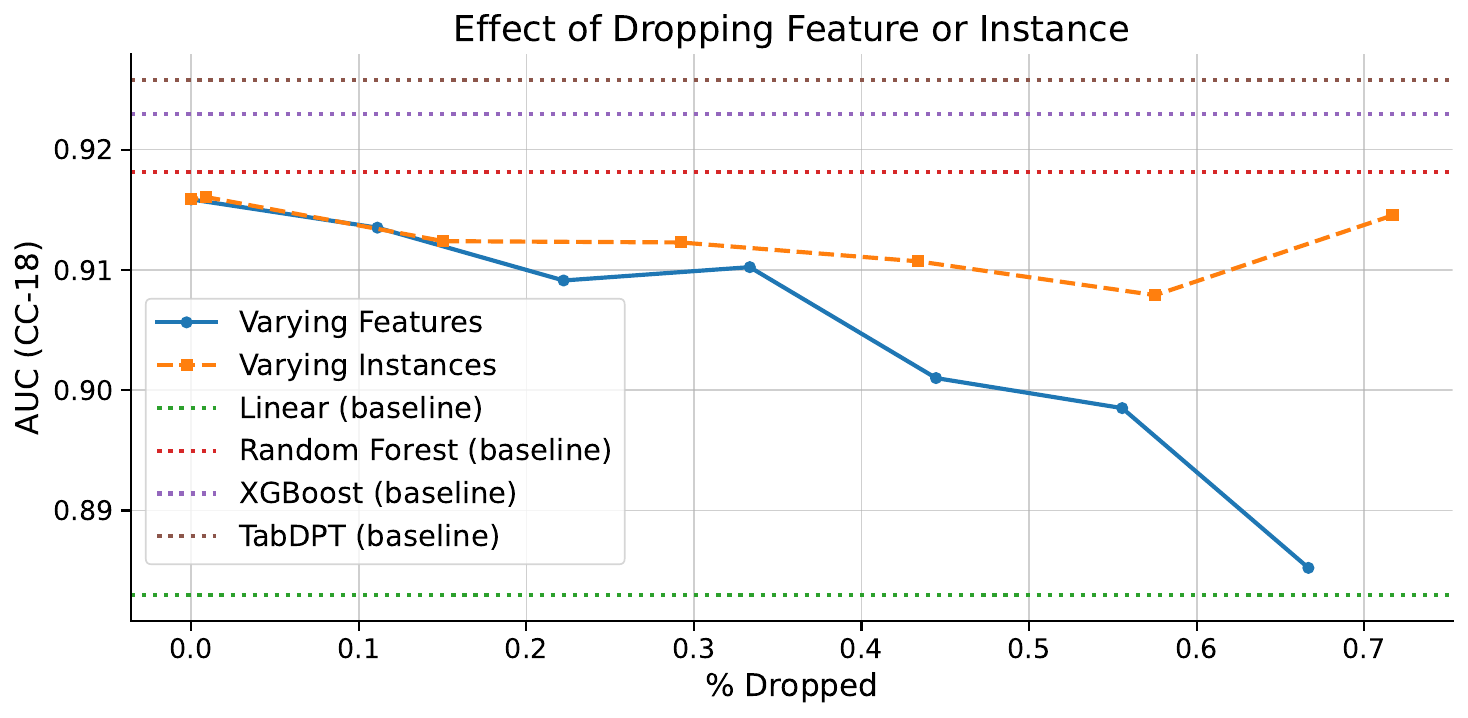}
        \caption{Column vs.\ Row Importance: We analyze the downstream performance when removing either features or instances from the \texttt{COLLEGES} dataset. 
        }
        \label{fig:feature_instance}
    \end{subfigure}
    \caption{Not all cells in a table are made equal: the number of features matters much more than the number of instances as pre-training dataset for tabular ICL models.}
    \vspace{-1em}
    \label{fig:universal1}
\end{figure}

\looseness=-1 Using these meta-features, we train an XGBoost~\citep{chen2016xgboost} regressor to predict a dataset's average downstream generalization score, defined as the average of correlation for regression tasks and accuracy for classification tasks.
The resulting model achieves an $R^2$ of $0.67$, indicating that roughly two-thirds of the variance in generalization can be reproduced from straightforward dataset properties alone. 
Notably, when evaluating feature importance (see~\Cref{fig:xgb_feature_importance}), the number of features emerges as by far the strongest predictor of downstream transfer performance.
In contrast, the number of instances---often presumed critical in classical settings~\citep{vapnik2013nature}---shows negligible predictive power.
This suggests that the richness of the feature space is far more important than purely the dataset size for tabular ICL-based generalization. We validate this in~\Cref{fig:feature_instance} by removing a fraction of either rows or columns on the \textsc{Colleges} dataset, showing that features tend to matter much more than instances. After dropping about $70\%$ of features, the model performance drops to a similar level as a linear model, while dropping more than $70\%$ of instances still allows the model to maintain performance close to a random forest. 
We see a similar phenomenon across many other datasets. %

\subsection{Training on many good tasks unlocks generalization}

\begin{wrapfigure}{r}{0.52\textwidth} %
  \vspace{-2.5em}             %
  \centering
  \includegraphics[width=\linewidth]{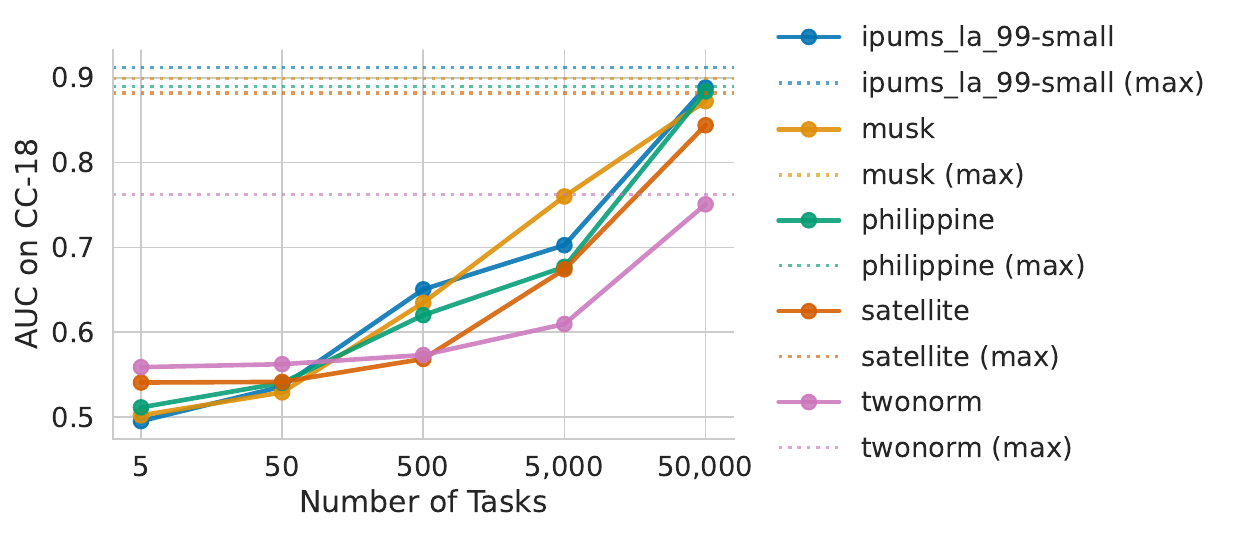}
  \caption{\looseness=-1 Downstream AUC vs.\ the number of
  unique tasks used during training. 
  \texttt{(max)} denotes the unrestricted setting where the sampler draws from the full set of eligible task configurations with no upper count limit.
  More tasks during pre-training
  consistently leads to better transfer. %
  }
  \label{fig:num_tasks_performance}
  \vspace{-1.5em}             %
\end{wrapfigure}

\looseness=-1 To understand why dropping features caused a strong performance drop, we revisit the randomized procedure used for pre-training: a column is selected as target while a subset of other columns are used as features; we call this a \emph{task}. Using a fixed supervised target provides just one task, while using a randomized target and features can have $\mathcal{O}(k\ 2^k)$ tasks for $k$ columns. Thus, the role of columns may be more important than the role of rows as it is directly tied to the number of tasks or feature relationships the model is exposed to during training.

\looseness=-1
We disentangle the impact of the number of tasks from the number of features in~\Cref{fig:num_tasks_performance}. We do not change the amount of instances or features in a given dataset but instead we vary the number of tasks the model is allowed to pre-train on.
To be precise, we select $N$ tasks using allowed permutations (one target, the remaining columns being either masked or used as features) and vary $N$ for several datasets.
A striking pattern emerges: across all datasets, increasing the number of tasks from 5 to tens of thousands drives the performance from barely above random ($\text{AUC} = 0.5$) to good. Thus, although inherent dataset differences matter and some datasets are significantly better for pre-training, the main factor explaining cross-domain generalization is the number of tasks seen during pre-training.
In other words, the key is not the amount of pre-training data in tokens or bytes of memory, but the number of tasks/feature-target relationships that enable a tabular ICL model to generalize.

\looseness=-1
We further support the task-diversity view by training models on pairs with one dataset from the top quartile and one from the bottom quartile of Table~\ref{tab:cc18_auc}. Pairing a strong dataset with a weak one does not hurt performance and can even slightly improve it: \textsc{Colleges} ($\text{AUC}=0.918$ alone) with \textsc{Analcatdata\_Supreme} ($0.814$ alone) reaches $0.919$, and \textsc{APSFailure} ($0.918$ alone) with \textsc{Poker-Hand} ($0.777$ alone) also reaches $0.919$. These results suggest that datasets performing poorly during single-table pre-training are not intrinsically uninformative: they may be task-poor, but their tasks can complement those of a stronger dataset. This pattern also reappears in the next section.

\section{From one to many: pre-training insights for large corpora}
\label{sec:many_setup}

\looseness=-1
The preceding section showed how pre-training data properties such as feature count, sample size, 
and task diversity affect generalization in the single-dataset regime. We extend these observations to the practical setting where a TFM is pre-trained on a large, heterogeneous corpus of real-world tabular datasets.  We ask the question of how to curate and pre-process such a corpus.
We ablate along two axes: (i) \textbf{dataset-level selection:}
which datasets to include in the corpus, and (ii) \textbf{feature-level pre-processing:}
how to filter the features of each dataset prior to training. Our key findings are that exact deduplication is unnecessary, aggressive feature-based dataset-level filtering hurts performance, and lightweight column-level pre-processing is consistently beneficial.

\looseness=-1 We use part of the datasets collected by~\citet{hosseinzadeh2026tabdpt} as our pre-training corpus: $1{,}732$ datasets collected from OpenML~\citep{OpenML2025}, excluding datasets from the CC-18, CTR-23, and TabArena benchmarks~\citep{erickson2025tabarena}, which we use for evaluation.
We use a training context length of $1{,}024$ ($512$ queries) and an inference context length of $2{,}048$, and do not pre-process evaluation datasets. Our main evaluation metric is Interquartile Mean (IQM), which is standard and robust to outliers~\citep{agarwal2021deep, ma2025tabdptscalingtabularfoundation}. We defer the full experimental setup, including model and training details, to Appendix~\ref{app:training:base}.

\subsection{Does dataset-level selection help?}

\looseness=-1 Two natural corpus-curating strategies would be deduplication and filtering out ``universally bad'' datasets. We find that neither is beneficial. \textbf{Deduplication:} Although deduplication benefits LLM training~\citep{lee-etal-2022-deduplicating}, exact deduplication is unnecessary here: models trained on the full corpus ($1{,}732$ datasets) and a deduplicated corpus ($1{,}535$ datasets, identified by the SHA-256 hash of $X$ and $y$) are statistically indistinguishable on CC-18 and CTR-23 (\Cref{tab:corpus_ablation}). We attribute this to the SSL objective, which already reduces the impact of duplicates through random column-target sampling and independent context windows at each step. \textbf{Dataset-level filtering:} A reductive direction here would be to restrict the pre-training corpus to feature-rich datasets, since we established that those generalize better than low-feature ones. However, we previously showed that even a table with very few features can induce non-trivial out-of-domain generalization. Low-feature datasets are qualitatively different from high-feature ones, covering simpler dependency structures, smaller column-count regimes, and distinct domains. We test this by restricting both the full and deduplicated corpus to datasets with at least $30$ features ($607$ datasets, and $557$ after deduplication). \Cref{tab:corpus_ablation} shows that this degrades performance across the board, with similar performance regardless of deduplication: excluding low-feature datasets reduces the number of diverse tasks in a way not offset by higher average per-dataset quality.

\subsection{Column-level pre-processing as a better alternative}

\looseness=-1 Since coarse dataset filtering is not fruitful, we instead perform fine-grained feature cleaning to increase task quality and diversity. For each dataset, we apply a subset of three operations, each ablated individually by training a separate model for every combination. \textbf{NaN-column drop} removes columns where more than $50\%$ of values are missing; after TabDPT's mean imputation, such columns become nearly constant, so dropping them improves task quality. \textbf{Correlation deduplication} greedily removes columns until no pair has absolute correlation above $0.90$, improving task diversity. The target $y$ is included in the joint correlation matrix so features near-collinear with it are also removed, while $y$ itself is protected from removal. We test Pearson and Spearman variants. At each greedy step, the column with the greatest number of above-threshold correlation partners is removed, with ties broken by lower variance. All columns are treated as numeric: categorical features are pre-encoded as integers and used as-is. \textbf{Minimum-feature enforcement} (mf-$k$) skips any dataset for which fewer than $k$ feature columns remain after the above steps; we set $k=5$ by default. Unlike the two column operations above, mf-$k$ is a dataset-level exclusion rule rather than a column transformation: it is applied \emph{after} column cleaning to drop tables left with too few usable columns. It's a light cleaning step that preserves most of the corpus rather than an aggressive dataset selection rule that discards whole task pools.

\begin{table}[t]
\centering
\vspace{-1em}
\caption{\looseness=-1 Dataset-level selection and column-level pre-processing ablations. Results are \textbf{IQM} with $95\%$ CI; AUC/accuracy on CC-18, correlation/$R^2$ on CTR-23, Avg.\ is unweighted. First row is the full-corpus baseline; dataset-level rows test deduplication, $\geq\!30$ feature filtering, and both. Column-level \textbf{Setup} = (NaN, Corr, mf-$5$): `+' enabled, `--' disabled; NaN drops columns with $>50\%$ missing, Corr is Pearson (P) or Spearman (S) deduplication at threshold $0.90$, and mf-$5$ skips datasets with $<5$ surviving features. Best column-level IQM per metric is \textbf{bold}.}
\label{tab:corpus_ablation}
\setlength{\tabcolsep}{2pt}
\scalebox{1.0}{
\begin{tabular}{lccccc}
\toprule
\textbf{Setup} & \textbf{AUC} & \textbf{Acc.} & \textbf{Corr.} & $\boldsymbol{R^2}$ & \textbf{Avg.} \\
\midrule
Baseline
    & ${0.9757}_{\scriptscriptstyle[.9731,.9776]}$
    & ${0.9233}_{\scriptscriptstyle[.9199,.9265]}$
    & ${0.9021}_{\scriptscriptstyle[.8993,.9047]}$
    & ${0.8145}_{\scriptscriptstyle[.8097,.8188]}$
    & $0.9039$ \\
\midrule
\multicolumn{6}{l}{\textit{Dataset-level selection}} \\
Deduped
    & ${0.9752}_{\scriptscriptstyle[.9728,.9773]}$
    & ${0.9236}_{\scriptscriptstyle[.9202,.9269]}$
    & ${0.9019}_{\scriptscriptstyle[.8989,.9047]}$
    & ${0.8146}_{\scriptscriptstyle[.8101,.8189]}$
    & $0.9038$ \\
$\geq\!30$
    & ${0.9728}_{\scriptscriptstyle[.9704,.9748]}$
    & ${0.9208}_{\scriptscriptstyle[.9175,.9241]}$
    & ${0.8882}_{\scriptscriptstyle[.8832,.8930]}$
    & ${0.7890}_{\scriptscriptstyle[.7812,.7966]}$
    & $0.8927$ \\
$\geq\!30$ + dedup
    & ${0.9724}_{\scriptscriptstyle[.9701,.9744]}$
    & ${0.9195}_{\scriptscriptstyle[.9159,.9231]}$
    & ${0.8871}_{\scriptscriptstyle[.8817,.8921]}$
    & ${0.7870}_{\scriptscriptstyle[.7788,.7949]}$
    & $0.8915$ \\
\midrule
\multicolumn{6}{l}{\textit{Column-level pre-processing}} \\
(+, --, --)
    & ${0.9760}_{\scriptscriptstyle[.9734,.9779]}$
    & ${0.9256}_{\scriptscriptstyle[.9226,.9286]}$
    & ${0.9078}_{\scriptscriptstyle[.9049,.9105]}$
    & ${0.8221}_{\scriptscriptstyle[.8177,.8261]}$
    & $0.9079$ \\
(--, P, --)
    & ${0.9760}_{\scriptscriptstyle[.9737,.9780]}$
    & ${0.9251}_{\scriptscriptstyle[.9222,.9280]}$
    & ${0.9043}_{\scriptscriptstyle[.9019,.9065]}$
    & ${0.8195}_{\scriptscriptstyle[.8157,.8230]}$
    & $0.9062$ \\
(--, S, --)
    & ${0.9758}_{\scriptscriptstyle[.9734,.9778]}$
    & ${0.9238}_{\scriptscriptstyle[.9209,.9267]}$
    & ${0.9032}_{\scriptscriptstyle[.9006,.9055]}$
    & ${0.8164}_{\scriptscriptstyle[.8124,.8201]}$
    & $0.9048$ \\
(+, --, +)
    & $\mathbf{0.9763}_{\scriptscriptstyle[.9737,.9783]}$
    & ${0.9251}_{\scriptscriptstyle[.9221,.9280]}$
    & ${0.9079}_{\scriptscriptstyle[.9053,.9103]}$
    & ${0.8249}_{\scriptscriptstyle[.8207,.8289]}$
    & $0.9085$ \\
(+, P, --)
    & ${0.9756}_{\scriptscriptstyle[.9732,.9777]}$
    & ${0.9239}_{\scriptscriptstyle[.9204,.9272]}$
    & ${0.9027}_{\scriptscriptstyle[.9002,.9050]}$
    & ${0.8166}_{\scriptscriptstyle[.8121,.8205]}$
    & $0.9047$ \\
(--, P, +)
    & ${0.9759}_{\scriptscriptstyle[.9734,.9778]}$
    & ${0.9250}_{\scriptscriptstyle[.9220,.9280]}$
    & $\mathbf{0.9098}_{\scriptscriptstyle[.9071,.9125]}$
    & $\mathbf{0.8281}_{\scriptscriptstyle[.8236,.8323]}$
    & $\mathbf{0.9097}$ \\
(+, P, +)
    & ${0.9756}_{\scriptscriptstyle[.9731,.9775]}$
    & ${0.9248}_{\scriptscriptstyle[.9219,.9277]}$
    & ${0.9060}_{\scriptscriptstyle[.9034,.9083]}$
    & ${0.8227}_{\scriptscriptstyle[.8186,.8264]}$
    & $0.9073$ \\
(+, S, --)
    & ${0.9758}_{\scriptscriptstyle[.9734,.9778]}$
    & ${0.9241}_{\scriptscriptstyle[.9215,.9267]}$
    & ${0.9057}_{\scriptscriptstyle[.9031,.9082]}$
    & ${0.8203}_{\scriptscriptstyle[.8161,.8241]}$
    & $0.9065$ \\
(--, S, +)
    & ${0.9762}_{\scriptscriptstyle[.9737,.9782]}$
    & ${0.9255}_{\scriptscriptstyle[.9228,.9282]}$
    & ${0.9078}_{\scriptscriptstyle[.9053,.9102]}$
    & ${0.8241}_{\scriptscriptstyle[.8203,.8278]}$
    & $0.9084$ \\
(+, S, +)
    & $\mathbf{0.9763}_{\scriptscriptstyle[.9738,.9784]}$
    & $\mathbf{0.9272}_{\scriptscriptstyle[.9241,.9302]}$
    & ${0.9077}_{\scriptscriptstyle[.9055,.9097]}$
    & ${0.8251}_{\scriptscriptstyle[.8213,.8284]}$
    & $0.9091$ \\
\bottomrule
\end{tabular}}
\end{table}

\looseness=-1 \Cref{tab:corpus_ablation} reports the full ablation. 
Every pre-processing variant outperforms the baseline on regression, with IQM $R^2$ gains from $+0.002$ to $+0.014$. Classification gains are smaller but broadly consistent: all variants improve accuracy, and most improve AUC. This is consistent with the single-dataset findings: generalization is governed by the number of unique, good tasks the model is pre-trained on. Dataset filtering discards entire task pools, whereas column pre-processing removes only redundant or uninformative tasks without shrinking the corpus, thus increasing task diversity and quality. 

\subsection{Does this extend to larger models, context windows, and other evaluation benchmarks?}
\label{sec:larger}

\looseness=-1
We next ask whether the gains from column-level pre-processing generalize beyond the base model size, training context, and evaluation setting. To this end, we scale up along two axes simultaneously: model capacity (depth and width) and training context length. We train a deeper, wider model
under three maximum training context lengths: $1{,}024$ (``1k'', base), $2{,}048$ (``2k''), and $4{,}096$ (``4k'') tokens. Each variant is trained with and without pre-processing (we choose NaN\,$+$\,Spearman\,$+$\,mf-$5$, denoted ``$+$pp''). 
Inference continues to use context size $2{,}048$ and no pre-processing for consistency. Full architectural, training, and context size details for all six variants are listed in Appendix~\ref{app:training:large}.

\begin{figure}[h]
    \centering
    \includegraphics[width=.8\linewidth]{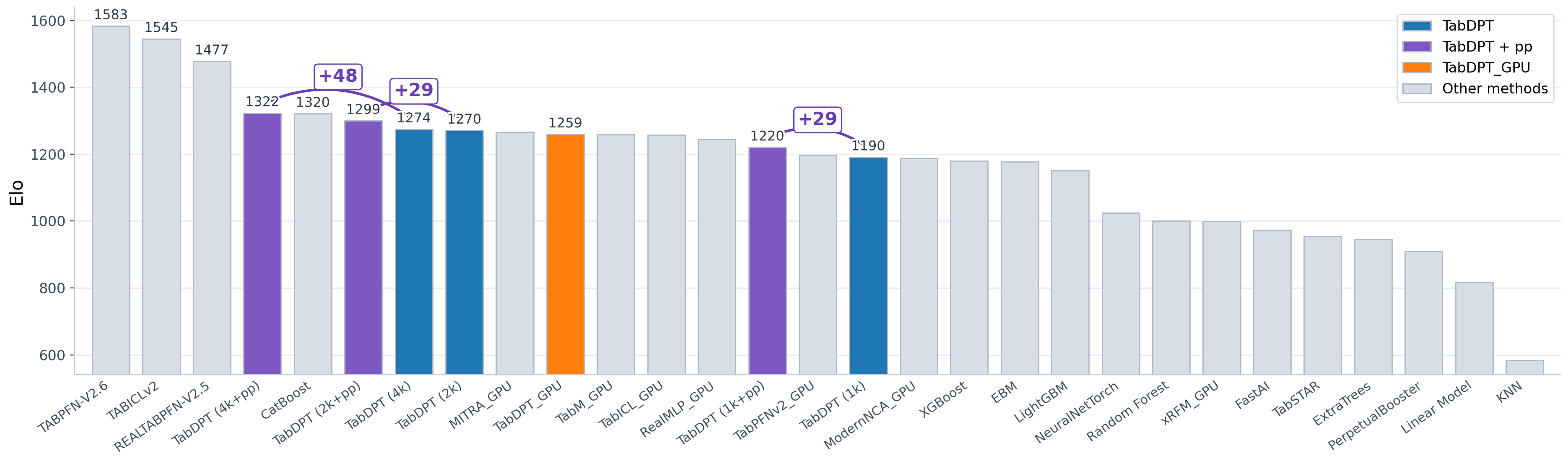}
    \caption{ %
    Pre-processing yields Elo gains at all context sizes.
    Purple bars are models trained with pre-processing (+pp), blue bars their corresponding baselines, and grey bars other TabArena methods. TabDPT~\citep{ma2025tabdptscalingtabularfoundation} is in orange. 
    }
    \label{fig:elo}
\end{figure}

\looseness=-1
We evaluate on TabArena's $51$ tabular prediction tasks (Apache License 2.0)~\citep{erickson2025tabarena}. We compute pairwise win rates across all configurations. The results are directionally consistent with our findings: the pre-processed variant outperforms its baseline counterpart at every context scale: 4k$+$pp beats 4k on $67\%$ of tasks, 2k$+$pp beats 2k on $61\%$, and 1k$+$pp beats 1k on $63\%$. The strongest overall configuration is 4k$+$pp.
We contextualize these findings within the broader TabArena leaderboard in~\Cref{fig:elo}, reporting Elo against some other default-configuration baselines, with details in Appendix~\ref{app:tabarena}.

\section{Understanding what makes generalization possible}
\label{sec:generalization}

\looseness=-1
The previous sections established two key empirical facts: first, that broad transfer can emerge from pre-training on a single table, and second, that this behaviour persists and can be systematically shaped when scaling to large corpora. These observations cannot be easily explained as parametric transfer learning where predictive signals from one dataset transfer to a related one. Instead, the model must have learned a general prediction algorithm, and we argue that tabular ICL models specifically rely on a \emph{learned retrieval-and-aggregation procedure}. In that sense, their form of generalization is closer to that of $k$-nearest neighbours than a standard fitted predictor---indeed, it would not be surprising for a non-parametric method such as $k$NN to perform reasonably well on both an \textsc{MNIST}-like task and a \textsc{California-Housing}-like task. The role of pre-training is thus not to store a universal decision rule in the weights, but rather to learn how to compare examples, identify context points relevant to a query, and aggregate their label information.

\looseness=-1
This retrieval-based perspective stands in contrast to previous works on prior-fitted networks, which interpret TFM pre-training as inducing a prior that corresponds to the space of processes that generate pre-training datasets~\cite{hollmann2023tabpfn,muller2022transformers}. In~\Cref{sec:single-table}, however, we observed that TFMs are effective on data outside of their training distribution, transferring between entirely different tabular datasets, and furthermore are effective when pre-trained with a very limited distribution.
The prior-fitting account still does not explain the case where pre-training and inference distributions have no overlap, which motivates our alternative retrieval-and-aggregation perspective of model behaviour.
Concurrent work casts further doubt on the prior-fitting account by showing that the predictions of current TFMs are neither marginalization- nor factorization-consistent, and therefore cannot be the conditionals of any single joint distribution over a table's columns~\cite{klotergens2026tabular}. We concede that approximate Bayesian inference could be occurring in models trained with broader distributions, but we believe that learned retrieval is a sufficient explanation for these models' capabilities, while being both more parsimonious and consistent with our observations.

\looseness=-1
In this section, we provide evidence for learned retrieval through three controlled analyses.
First, we show that models that generalize well are also better at retrieving exact label information from the context.
Second, we show that forcing the model to behave in a more explicit similarity-based way does not hurt performance, and can even help in weaker regimes.
Finally, we show that strong models tend to converge toward similar retrieval patterns, suggesting that there exists a fairly universal structure underlying good tabular generalization.

\subsection{Generalization is tied to the ability to retrieve from the context}

\begin{wrapfigure}{r}{0.35\linewidth}
    \vspace{-2em}
    \centering    \includegraphics[width=\linewidth]{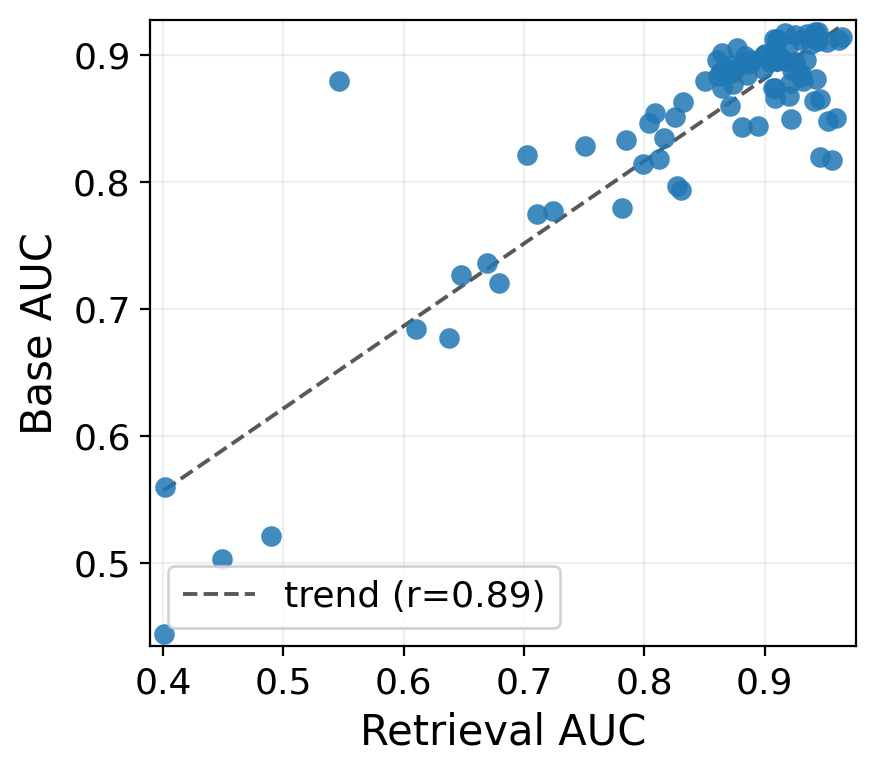}
    \caption{In-context retrieval AUC (x-axis) vs.\ base AUC (y-axis) for $88$ models each pre-trained on one dataset. Each point is one model.}
    \label{fig:retrieval_head8}
    \vspace{-1em}
\end{wrapfigure}

\looseness=-1
Based on our interpretation, the ability to generalize to new datasets should be tightly linked to a more primitive capability: the ability to identify and use the relevant information already present in the context.
A model that cannot reliably retrieve useful in-context examples would have little hope of producing robust predictions on novel datasets.
Conversely, a model that does generalize well should also perform well on tasks where the solution is effectively already available in the context and the main challenge is to recover it.
To test this, we construct a controlled retrieval task.  
For each model pre-trained on a single table, for evaluation, we run standard ICL inference, but \textbf{with the query rows set to be the same as context rows}, effectively testing whether the model can correctly retrieve and match identical inputs.
\looseness=-1 Concretely, we use the CC-18 classification datasets with a stratified subsample of up to $1024$ examples. The model is then given a sequence consisting of labelled examples followed by the same examples without labels. Performance is evaluated using AUC against the true labels, which we call \textbf{retrieval AUC}. %

\looseness=-1
Because every query row has an exact copy with its label in the context, a model that reliably attends to and copies the matching support label should score highly, although not perfectly: aggregation over several context rows can dilute or override the exact match, exactly as in $k$NN with $k>1$. A model that ignores the context and relies purely on parametric knowledge would revert to marginal or random predictions. The procedure is repeated with $n = 3$ independent stratified subsamples per dataset and the scores are averaged. We report the mean score across CC-18 datasets. The reported \textbf{base AUC} for each model is its held-out test-set AUC on CC-18 (standard ICL with distinct query and context rows) which we use as a proxy for model performance.
This highlights the \textit{retrieve and copy} skill as a core mechanism that tabular ICL must learn in order to generalize, similarly to the induction head mechanism in LLMs~\citep{olsson2022context}.

\looseness=-1
\pgraph{Results} Figure~\ref{fig:retrieval_head8} shows results for $88$ single-table pre-trained models. %
We find a strong correlation (Pearson  = $0.89$) between the standard generalization performance on held-out subsets and models' performance in this retrieval-oriented setting. Models that are weak at out-of-distribution generalization are also weak at exploiting the labelled context itself, whereas models that generalize well are precisely those that are able to recover the relevant support information. This supports the view that tabular ICL is not primarily driven by memorizing a broad family of parametric predictors in the weights. Instead, a substantial part of the problem appears to consist of learning a robust similarity-based procedure for retrieving and aggregating information from the context. We stress that exact-copy retrieval and downstream performance are distinct capabilities, measured under different prediction protocols, so the two AUCs are not comparable in absolute terms. The informative signal is how they vary together.

\subsection{Forcing a retrieval-based attention mechanism improves performance for weaker models}

\looseness=-1 If good generalization relies on learned retrieval, then one should expect tabular ICL models to be relatively robust to interventions that make them behave more explicitly like nearest-neighbour methods.
Specifically, if such models identify similar context points and aggregate their labels, then constraining the attention mechanism to a simpler similarity rule should not degrade performance, but actually improve it in regimes where the model fails to generalize, by providing a useful inductive structure.

\looseness=-1
We evaluate this by sharing the query and key projection weights for each attention layer, thus making each layer more ``$k$NN-like'' as the attention similarity score explicitly encodes a weighted inner product.
Indeed, as we use \textit{qk}-normalization, the attention score between points $x_i$ and $x_j$ when setting $W_Q = W_K$ becomes $q_i^\top q_j = \text{constant} - \frac{1}{2} \|q_i - q_j\|^2$.
This intervention strictly reduces the expressiveness of the attention mechanism---making it closer to a Gaussian $k$NN---and tests whether the model would only require a symmetric comparison operator in order to generalize.

\begin{figure}[h]
\centering
\begin{subfigure}[b]{\textwidth}
    \centering
    \includegraphics[width=.29\linewidth]{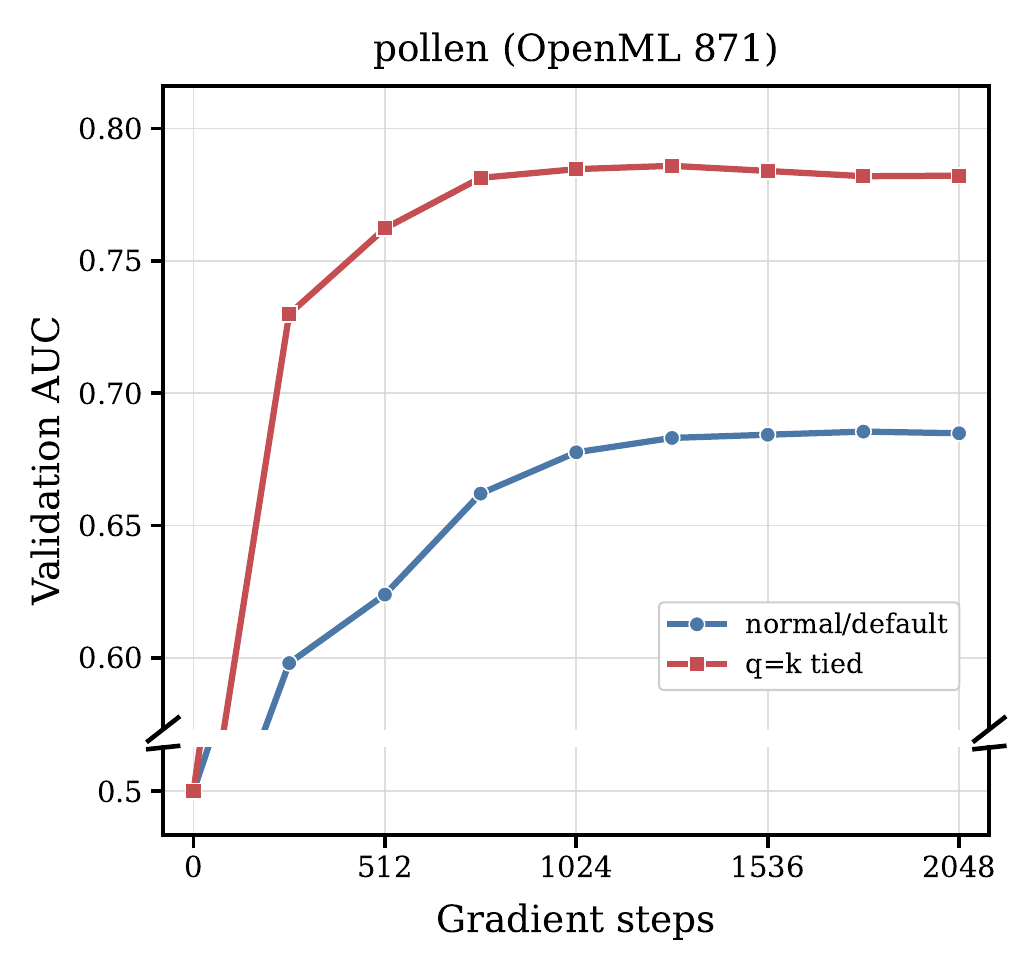}\hfill
    \includegraphics[width=.29\linewidth]{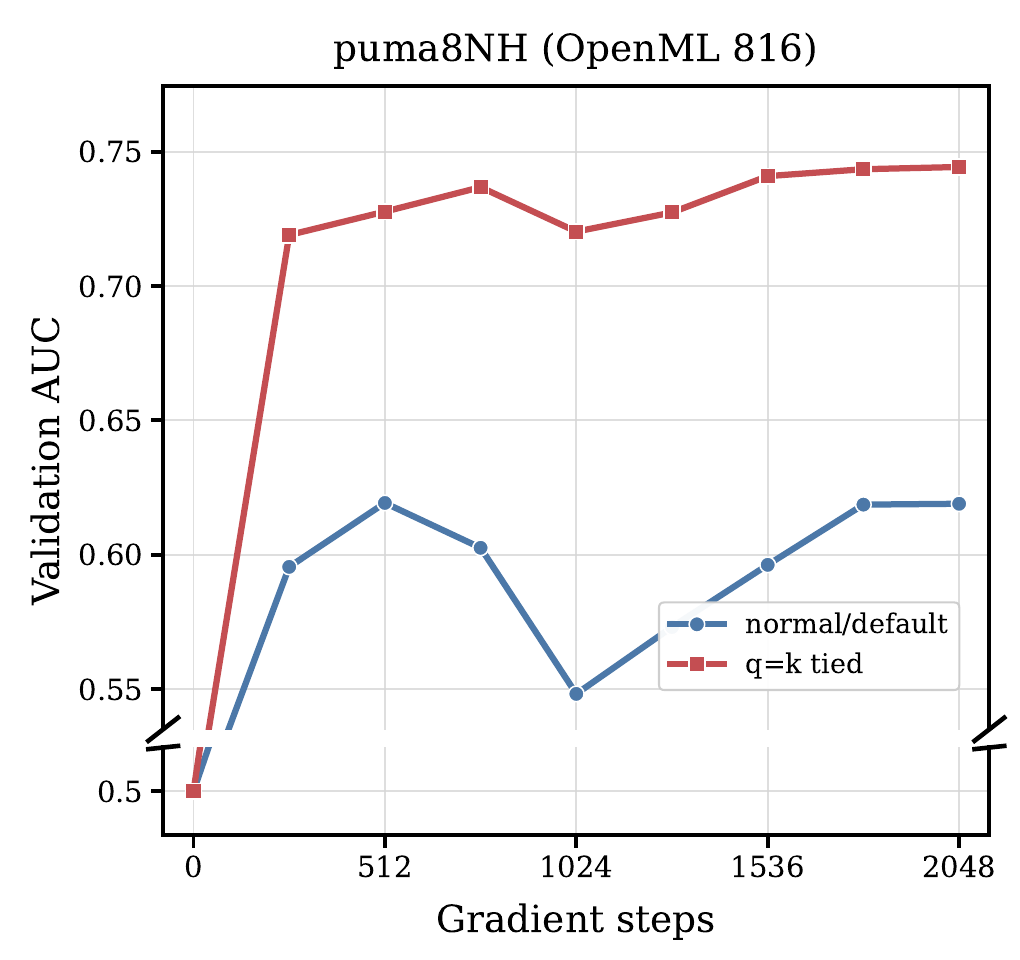}\hfill
    \includegraphics[width=.29\linewidth]{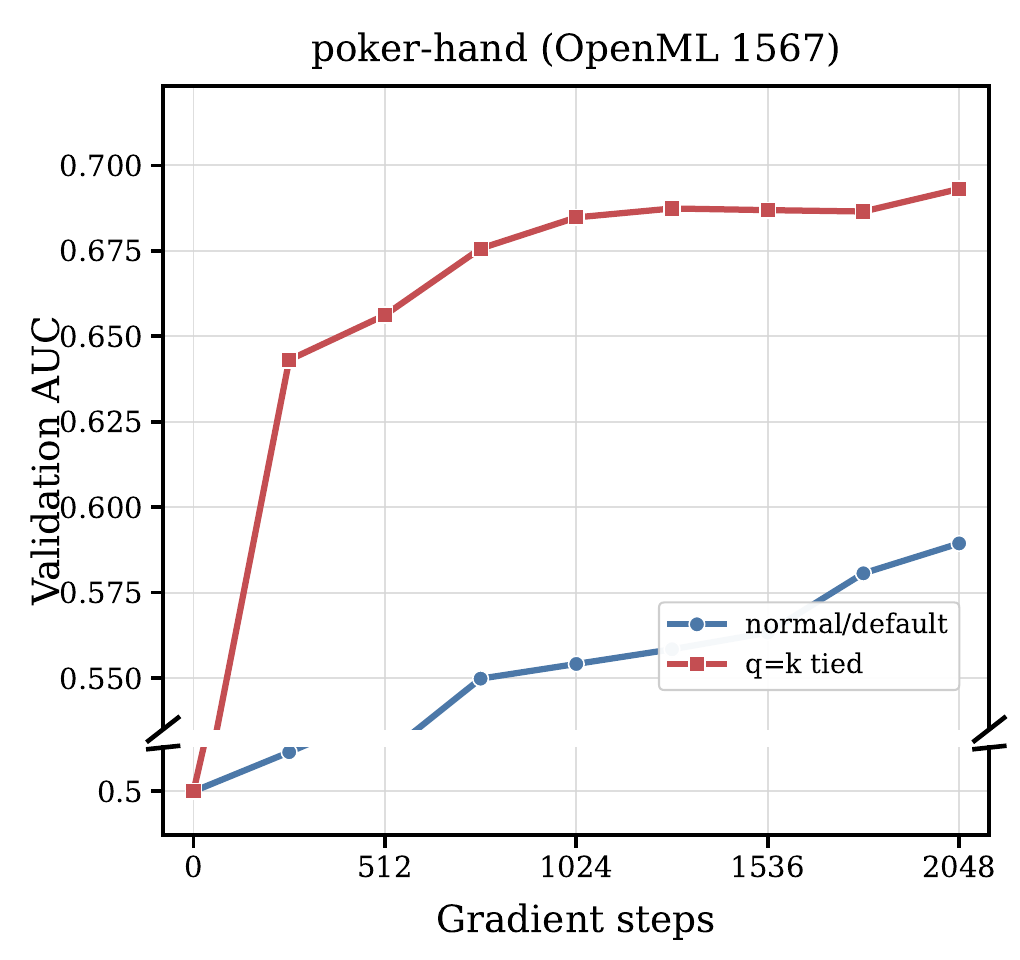}
    \caption{Tables that generalize poorly: \texttt{pollen}, \texttt{puma8NH}, \texttt{poker-hand} (left to right).}
    \label{fig:qk_ablation_weak}
\end{subfigure}

\begin{subfigure}[b]{\textwidth}
    \centering
    \includegraphics[width=.29\linewidth]{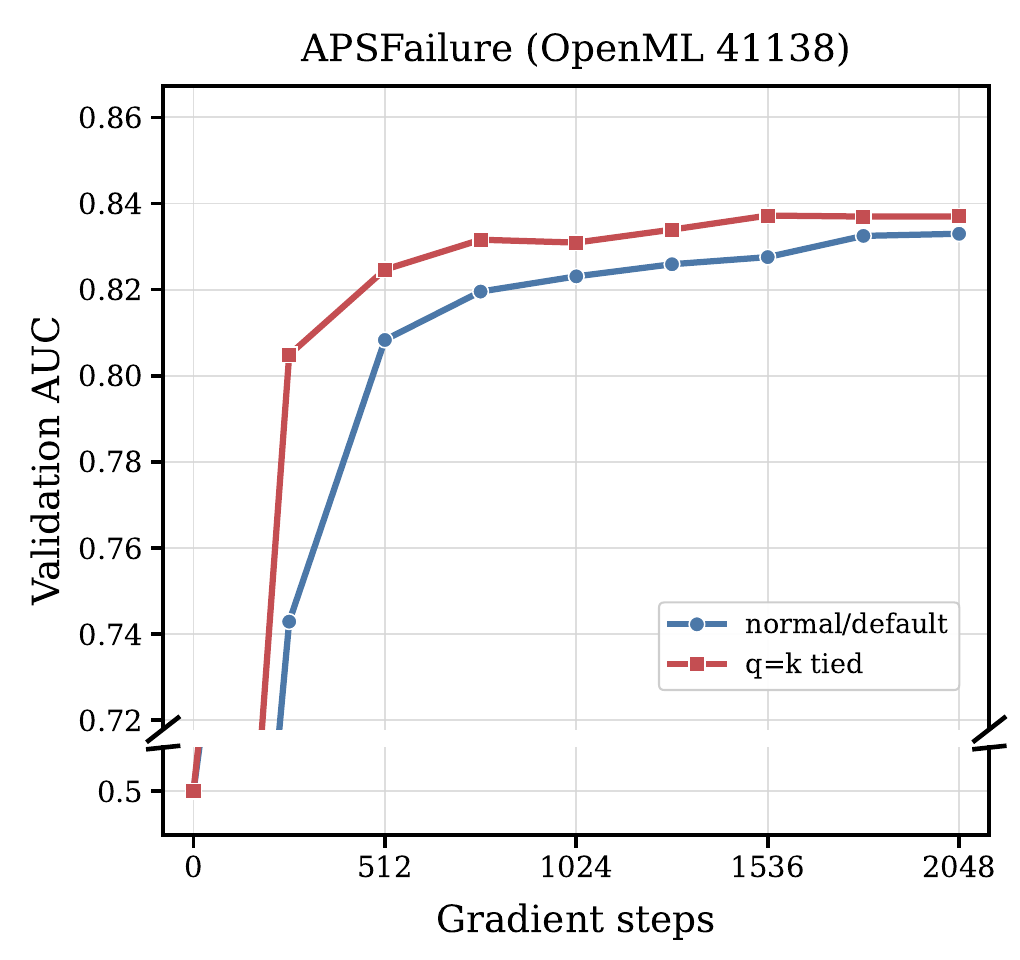}\hfill
    \includegraphics[width=.29\linewidth]{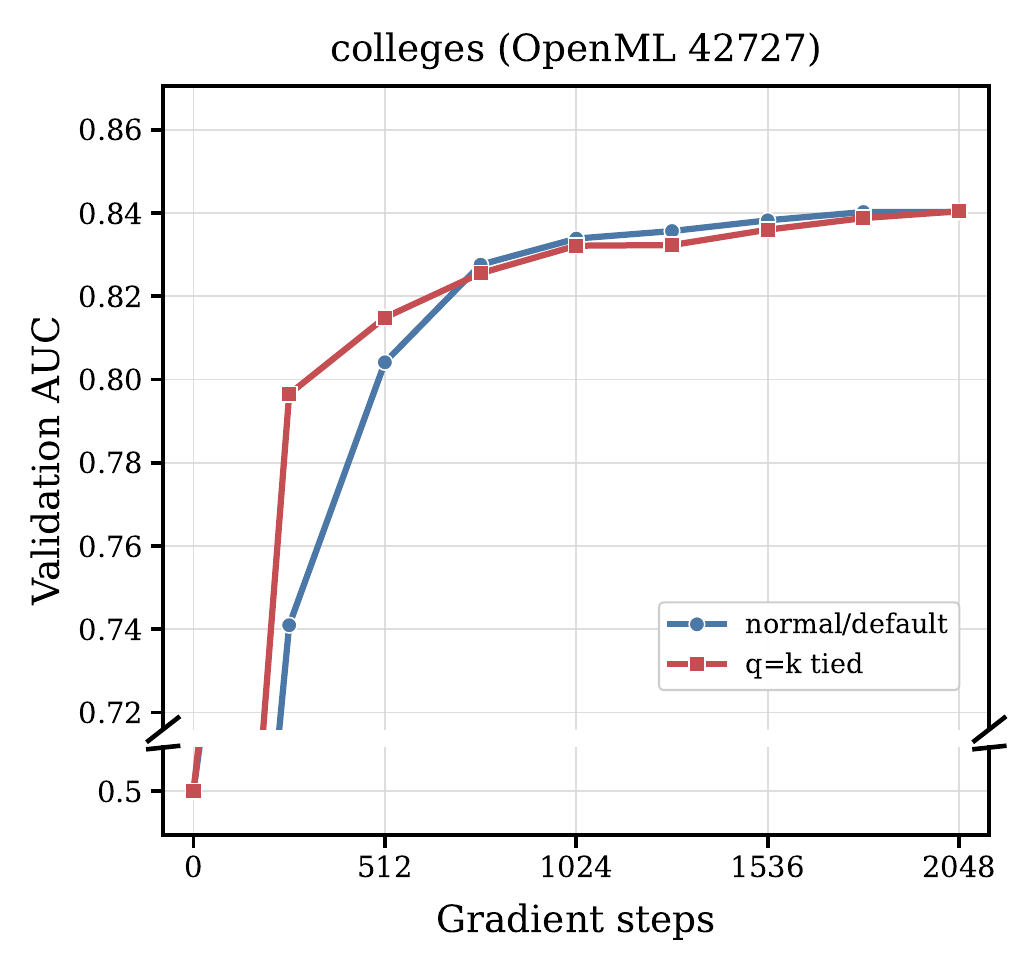}\hfill
    \includegraphics[width=.29\linewidth]{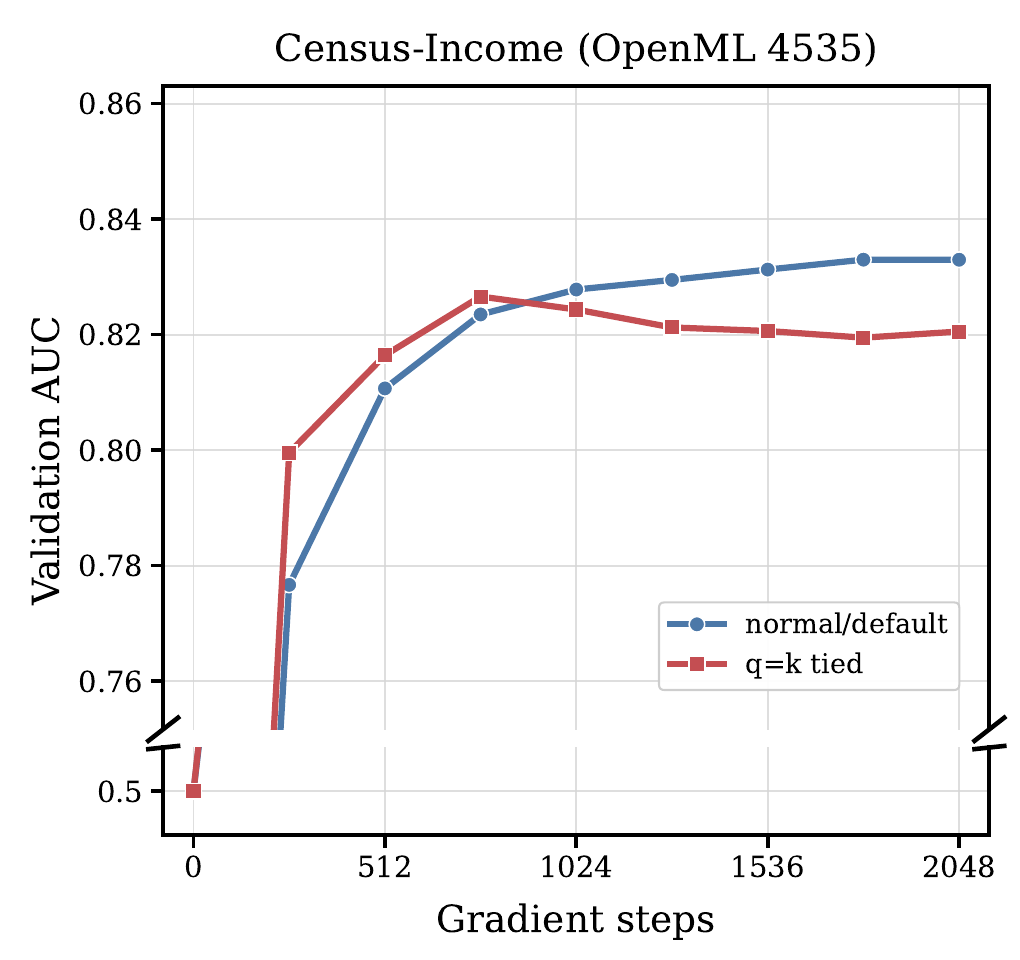}
    \caption{Tables that already generalize well: \texttt{APSFailure}, \texttt{Colleges}, \texttt{Census-Income} (left to right).}
    \label{fig:qk_ablation_strong}
\end{subfigure}
\caption{\looseness=-1 Ablation forcing $W_Q=W_K$ on six single-table pre-training corpora, grouped by how well the default architecture generalizes with single-table pre-training.}
\label{fig:qk_ablation}
\end{figure}

\begin{wrapfigure}{r}{0.39\textwidth}
\vspace{-2mm}
\centering
    \includegraphics[width=\linewidth]{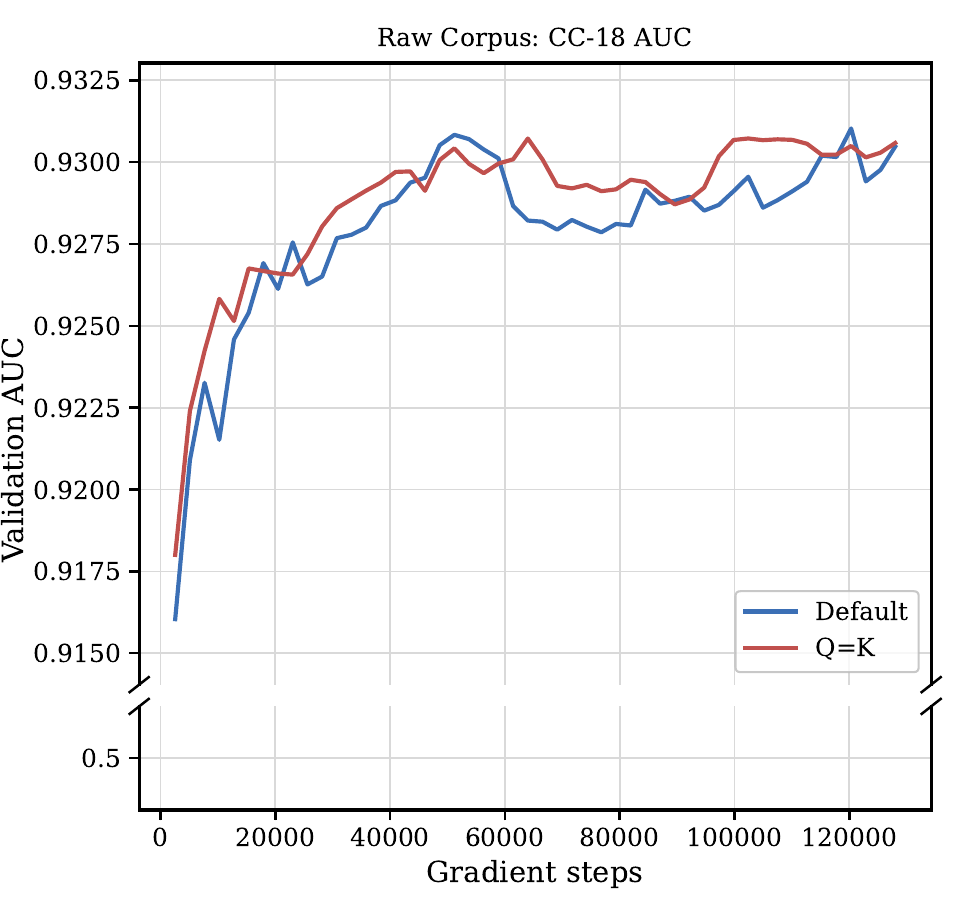}
\caption{Mean CC-18 AUC for the default architecture and the $W_Q=W_K$ variant, both pre-trained on the full OpenML corpus of \Cref{sec:many_setup}.}
\label{fig:qk_corpus}
\vspace{-4mm}
\end{wrapfigure}

\looseness=-1 We first target pre-training tables that generalize \emph{poorly}, \emph{i.e.}, tables whose resulting models attain low downstream AUC across the evaluation suite (selected using \Cref{tab:cc18_auc}).
This is where (i) the default architecture struggles to exploit labelled context examples from the pre-training table alone, and (ii) where the $W_Q = W_K$ constraint should help most under the retrieval interpretation.
In \Cref{fig:qk_ablation_weak}, the simplified version wins on all three such tables (\texttt{pollen}, \texttt{puma8NH}, \texttt{poker-hand}), by $+0.1088$ AUC on average, with analogous results on two further weak tables in Appendix~\ref{app:qk_extra}.
On three tables that already generalize well (\texttt{APSFailure}, \texttt{Colleges}, \texttt{Census-Income}), where an imposed similarity rule has little left to contribute, the constraint gives no systematic gain but also no substantial degradation, with a mean difference of $-0.0027$ AUC (\Cref{fig:qk_ablation_strong}).
Sharing the query and key projections therefore supplies a retrieval-inducing bias that matters when the pre-training table is too weak to teach a stable comparison rule, and becomes redundant, or at most mildly restrictive, once it is not. Finally, we scale the intervention to the full OpenML pre-training corpus used in \Cref{sec:many_setup}, where the two architectures are effectively tied: the constraint changes AUC by $+0.0001$ (\Cref{fig:qk_corpus}). We do not claim that $W_Q = W_K$ is a better architecture, as this gap is negligible.
The relevant point is that reducing attention to a strictly symmetric similarity rule has no noticeable cost even given abundant and diverse pre-training data, given the model sizes we used in our experiments.
Along with the weak-table gains, this reinforces the interpretation of tabular ICL as mostly retrieval-based in this setting, since distances between tokens under $W_Q = W_K$ are symmetric and analogous to soft-$k$NN.

\subsection{Strong models converge toward similar retrieval patterns}

\begin{wrapfigure}{l}{0.4\textwidth}
    \centering
    \vspace{-5mm}
    \includegraphics[width=\linewidth]{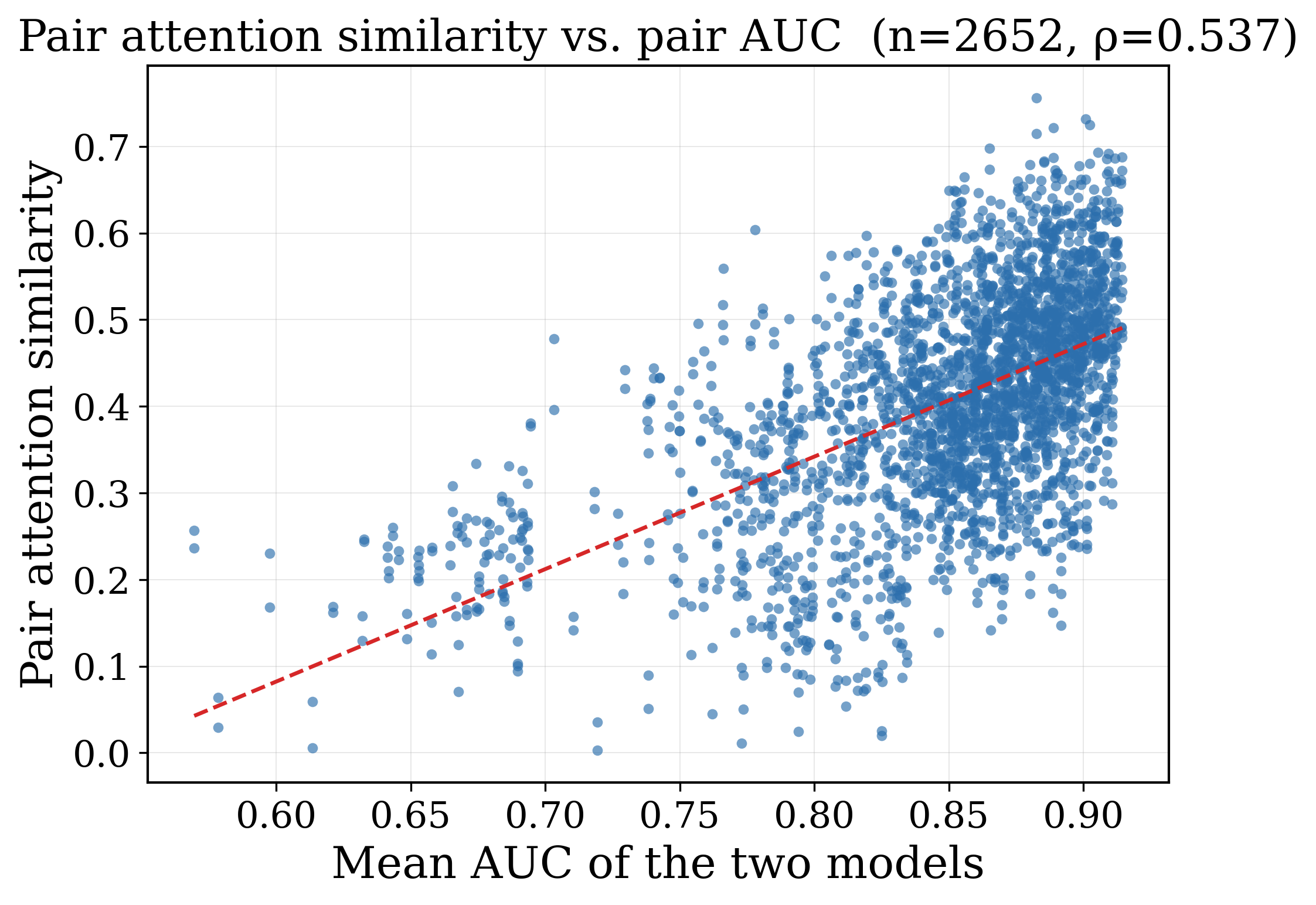}
    \caption{Pair attention similarity vs.\ mean AUC, $n=2652$ points (one per model pair $\times$ probe dataset), averaged over blocks. Spearman correlation is $0.537$. Per-block results are in Figure~\ref{fig:pair-auc-attn-bylayer}.}
    \label{fig:pair-auc-attn-overall}
    \vspace{-5mm}
\end{wrapfigure}

\looseness=-1
We further show that successful models should not merely perform well, but should do so in similar ways. 
If there exists a fairly universal retrieval strategy that works across tabular problems, then independently trained strong models should tend to converge toward comparable attention patterns. By contrast, if performance arose from brittle or highly idiosyncratic solutions, one would expect much more diversity across models. %

\looseness=-1 We test this prediction directly on $52$ TabDPT checkpoints each trained on a single OpenML dataset (single attention head per block for simplicity, %
model and training details are in Appendix~\ref{app:attn}). We probe each model on two further held-out OpenML datasets via forward passes to extract attention maps. For each (model, probe), we record the per-block, per-query attention distribution at a fixed in-context cutoff: the softmax-normalized vector a query token assigns over the $C$ context tokens. We compare two models row-by-row using Spearman rank correlation, averaging over queries and probe examples to obtain one similarity scalar per (model pair, block, probe) and then average the resulting scalar over the 12 blocks, yielding one pair-level value per (model pair, probe). Across them, the Spearman correlation between mean held-out AUC and pair attention similarity is $0.537$ (Figure~\ref{fig:pair-auc-attn-overall}).

\begin{wraptable}{r}{0.58\linewidth}
  \centering
  \vspace{-5mm}
  \caption{Pair attention similarity by category. CIs are non-parametric bootstrap intervals ($10{,}000$ resamples) on the difference of group means.}
  \resizebox{\linewidth}{!}{
  \label{tab:attn-summary}
  \begin{tabular}{lccccc}
    \toprule
    Group & Mean &  $n$ & vs.\ & $\Delta$ & $95\%$ CI \\
    \midrule
    SS & $\mathbf{0.504}$ & $156$ & SW & $+0.163$ & $[+0.143,+0.183]$ \\
    SW & $0.341$          & $338$ & WW & $+0.051$ & $[+0.027,+0.075]$ \\
    WW & $0.290$          &  $156$ &    &          &                   \\
    \bottomrule
  \end{tabular}}
  \vspace{-2em}
\end{wraptable}

To isolate what drives this correlation, we group models by performance, splitting by AUC quartile. We form three pair categories: strong--strong (SS, both endpoints in the top quartile%
), strong--weak (SW), and weak--weak (WW, both in the bottom quartile). 
The results of this experiment are in \Cref{tab:attn-summary} and the full methodological details (model specification, similarity definition, bootstrap-CI procedure, etc.) are in Appendix~\ref{app:attn}, which also contains per-block correlation values.

\looseness=-1
This convergence is unlikely to be accidental. Although checkpoints are trained on different datasets, high-AUC models independently discover similar attention patterns, as expected if good generalization follows a common retrieval principle. Together with the preceding analyses, these results suggest that tabular ICL models generalize not because they encode one gigantic universal predictor in their weights, but because they learn how to use the context at inference time. A simple similarity rule is often already enough to obtain non-trivial transfer, which explains why generalization is possible at all. The strongest models then improve upon this baseline by learning a more refined, more robust notion of relevance and aggregation.

\section{Related Work}

\looseness=-1
Although deep learning and gradient boosted tree ensemble algorithms are currently dominant on tabular benchmarks~\cite{erickson2025tabarena}, our work demonstrates a connection to \(k\)-nearest neighbours~\cite{fix1951discriminatory} and possibly kernel-based extensions~\cite{zuo2008kernel}. Explicit \(k\)-nearest retrieval has had recent applications as part of tabular ICL models~\cite{thomas2024retrieval,ma2025tabdptscalingtabularfoundation,zhang2025limix}.

\looseness=-1 While previous works on tabular ICL have explored many aspects of training task creation, they have generally focused on scaling the diversity of tasks, either through complex synthetic priors~\citep{hollmann2023tabpfn,Hollmann2025,qu2025tabicl,qu2026tabiclv2,zhang2025limix,zhang2025mitra} or augmented real world data~\citep{ma2025tabdptscalingtabularfoundation,garg2025real}, and the regime of extremely limited pre-training data has not been explored. Other surprising generalization capabilities of tabular ICL models under modified training approaches include resilience to distribution shifts~\cite{helli2024drift} and causal inference~\cite{balazadeh2025causal,robertson2025do}.

\looseness=-1 Substantial literature exists analyzing and extending in-context learning in large language models~\cite{dong2024survey}, however this is an emergent capability seen in autoregressive next-token prediction models trained on large amounts of text data, which is a fundamentally different setting from TFMs that directly train on an ICL objective. The observations in this paper are therefore separate from LLM ICL behaviour. While some LLM-focused papers have evaluated non-text data (e.g., \citep{chan2022data,reddy2023mechanistic,garg2022can}), they have still focused on autoregressive sequence modelling in ways that make them inapplicable to the models discussed here. Tabula-8B~\cite{gardner2024large} is an exception among tabular ICL models in that it uses a language model backbone, so we exclude it from our scope, but LLM-focused ICL findings could apply to it.

\section{Conclusion and limitations}
\label{sec:conclusion}

In this paper we investigated how pre-training data enables out-of-distribution generalization in TFMs. Taking a task-centric view, we explained why strong generalization can emerge even when training on a single real table, and expanded on this finding to make practical recommendations regarding dataset-level and column-level data filtering. Finally, we argued for a retrieval-focused interpretation of TFMs rather than one based on standard transfer learning or Bayesian inference, providing empirical evidence for retrieval behaviours. Overall, our findings challenge existing assumptions about how TFMs work and what actually matters when scaling TFM pre-training data.

\looseness=-1 Although we analyzed TFM behaviour across a broad range of datasets,  model sizes, and context lengths, we acknowledge other limitations in our work. We were unable to definitively identify or exclude additional mechanisms beyond learned retrieval that may underlie out-of-distribution generalization. Further research is therefore required to achieve a more comprehensive understanding of how in-context learning operates in TFMs. Moreover, our investigation was restricted to pre-training data derived from real-world datasets, and consequently, our findings may not fully extend to scenarios in which models are trained exclusively on synthetic data. Finally, while transformer architectures are universal across modern TFMs, they are split between row-based and cell-based attention, and our results only used row-based attention following TabDPT and TabPFNv1.

\section{Acknowledgements}

\looseness=-1 This work was supported by Mitacs through the Mitacs Accelerate program, and was enabled in part by compute resources provided by Mila (mila.quebec) and the Digital Research Alliance of Canada. We also thank Rasa Hosseinzadeh for the valuable discussions and ideas that contributed to this work.

\bibliography{refs}

\clearpage

\appendix

\section{Impact of pre-training dataset on CC-18 and CTR-23 Performance}

Tables~\ref{tab:cc18_auc} and ~\ref{tab:ctr23_r2} report the performance of TabDPT when pre-trained on different datasets and evaluated on two benchmark suites: \textbf{CC-18} for classification (measured by AUC) and \textbf{CTR-23} for regression (measured by $R^2$). Each table lists the results for all pre-training datasets, along with their number of instances, their number of features, and their domain. The results are sorted in descending order of performance, highlighting which pre-training datasets achieve better performance on each benchmark.

\begingroup
\setlength{\tabcolsep}{1pt}
\begin{longtable}{
@{}
>{\raggedright\arraybackslash}p{0.34\textwidth}
r
r
r
>{\raggedright\arraybackslash}p{0.24\textwidth}
@{}
}
\caption{AUC scores on CC-18 for TabDPT pre-trained on various datasets, sorted in descending order of AUC.}\label{tab:cc18_auc}\\
\toprule
\textbf{Dataset} & \textbf{AUC on CC-18} & \textbf{\# Instances} & \textbf{\# Features} & \textbf{Dataset Domain} \\
\midrule
\endfirsthead
\caption[]{AUC scores on CC-18 for TabDPT pre-trained on various datasets, sorted in descending order of AUC.}\\
\toprule
\textbf{Dataset} & \textbf{AUC on CC-18} & \textbf{\# Instances} & \textbf{\# Features} & \textbf{Dataset Domain} \\
\midrule
\endhead

\bottomrule
\endfoot
APSFailure & 0.918 & 76000 & 171 & industrial-operational \\
colleges & 0.918 & 7063 & 45 & other \\
ipums\_la\_99-small & 0.917 & 8844 & 57 & financial-demographic \\
ipums\_la\_97-small & 0.916 & 7019 & 61 & financial-demographic \\
ipums\_la\_98-small & 0.915 & 7485 & 56 & financial-demographic \\
dionis & 0.915 & 416188 & 61 & other \\
cardiotocography & 0.914 & 2126 & 36 & medical-human-sensor \\
Census-Income & 0.914 & 299285 & 42 & financial-demographic \\
volkert & 0.912 & 58310 & 181 & other \\
KDDCup09\_appetency & 0.912 & 50000 & 231 & human-behaviour \\
covertype & 0.912 & 581012 & 55 & biology-ecology \\
gas-drift-different-concentrations & 0.911 & 13910 & 130 & other-science \\
KDDCup09\_churn & 0.911 & 50000 & 231 & industrial-operational \\
road-safety & 0.910 & 111762 & 33 & human-behaviour \\
kick & 0.910 & 72983 & 33 & industrial-operational \\
eye\_movements & 0.908 & 10936 & 28 & medical-human-sensor \\
gas-drift & 0.906 & 13910 & 129 & other-science \\
philippine & 0.905 & 5832 & 309 & other \\
dilbert & 0.904 & 10000 & 2001 & other \\
christine & 0.901 & 5418 & 1637 & other \\
MiniBooNE & 0.900 & 130064 & 51 & physics-astronomy \\
Satellite & 0.900 & 5100 & 37 & physics-astronomy \\
jasmine & 0.899 & 2984 & 145 & other \\
scene & 0.899 & 2407 & 300 & vision-audio-text \\
helena & 0.896 & 65196 & 28 & other \\
kdd\_internet\_usage & 0.896 & 10108 & 69 & financial-demographic \\
SpeedDating & 0.895 & 8378 & 121 & human-behaviour \\
pol & 0.895 & 10082 & 27 & industrial-operational \\
elevators & 0.894 & 16599 & 19 & other \\
musk & 0.894 & 6598 & 168 & other-science \\
one-hundred-plants-texture & 0.894 & 1599 & 65 & biology-ecology \\
heloc & 0.894 & 10000 & 23 & financial-demographic \\
default-of-credit-card-clients & 0.893 & 13272 & 21 & financial-demographic \\
ada\_agnostic & 0.892 & 4562 & 49 & financial-demographic \\
sylva\_agnostic & 0.889 & 14395 & 217 & biology-ecology \\
colleges\_usnews & 0.889 & 1302 & 34 & other \\
spoken-arabic-digit & 0.887 & 263256 & 15 & vision-audio-text \\
ada & 0.887 & 4147 & 49 & other \\
jannis & 0.885 & 83733 & 55 & other \\
PizzaCutter3 & 0.885 & 1043 & 38 & other \\
fbis.wc & 0.884 & 2463 & 2001 & vision-audio-text \\
porto-seguro & 0.883 & 595212 & 58 & human-behaviour \\
sylvine & 0.883 & 5124 & 21 & other \\
cjs & 0.880 & 2796 & 35 & biology-ecology \\
Click\_prediction\_small & 0.879 & 39948 & 12 & human-behaviour \\
guillermo & 0.879 & 20000 & 4297 & other \\
mushroom & 0.879 & 8124 & 23 & biology-ecology \\
PieChart3 & 0.877 & 1077 & 38 & other \\
higgs & 0.877 & 98050 & 29 & physics-astronomy \\
magic & 0.874 & 19020 & 11 & physics-astronomy \\
ada\_prior & 0.874 & 4562 & 15 & financial-demographic \\
MagicTelescope & 0.873 & 19020 & 12 & physics-astronomy \\
eeg-eye-state & 0.867 & 14980 & 15 & medical-human-sensor \\
JapaneseVowels & 0.866 & 9961 & 15 & vision-audio-text \\
shuttle & 0.865 & 58000 & 10 & physics-astronomy \\
ldpa & 0.863 & 164860 & 8 & medical-human-sensor \\
okcupid-stem & 0.863 & 50789 & 20 & human-behaviour \\
house\_16H & 0.860 & 22784 & 17 & financial-demographic \\
credit & 0.854 & 16714 & 11 & financial-demographic \\
house\_8L & 0.851 & 22784 & 9 & financial-demographic \\
artificial-characters & 0.850 & 10218 & 8 & deterministic-simulated \\
page-blocks & 0.849 & 5473 & 11 & vision-audio-text \\
sf-police-incidents & 0.847 & 2215023 & 9 & human-behaviour \\
albert & 0.846 & 425240 & 79 & other \\
walking-activity & 0.844 & 149332 & 5 & medical-human-sensor \\
pbcseq & 0.843 & 1945 & 19 & medical-human-sensor \\
fabert & 0.834 & 8237 & 801 & other \\
compas-two-years & 0.833 & 4966 & 12 & human-behaviour \\
Diabetes130US & 0.828 & 71090 & 8 & medical-human-sensor \\
madeline & 0.821 & 3140 & 260 & other \\
airlines & 0.820 & 539383 & 8 & industrial-operational \\
analcatdata\_halloffame & 0.818 & 1340 & 17 & other \\
visualizing\_soil & 0.817 & 8641 & 5 & biology-ecology \\
analcatdata\_supreme & 0.814 & 4052 & 8 & other \\
chess & 0.796 & 28056 & 7 & deterministic-simulated \\
kropt & 0.794 & 28056 & 7 & deterministic-simulated \\
kr-vs-k & 0.780 & 28056 & 7 & deterministic-simulated \\
poker-hand & 0.777 & 1025009 & 11 & deterministic-simulated \\
pollen & 0.775 & 3848 & 6 & biology-ecology \\
puma8NH & 0.736 & 8192 & 9 & deterministic-simulated \\
yeast & 0.727 & 1269 & 9 & biology-ecology \\
hill-valley & 0.720 & 1212 & 101 & deterministic-simulated \\
imdb.drama & 0.684 & 120919 & 1002 & vision-audio-text \\
twonorm & 0.677 & 7400 & 21 & deterministic-simulated \\
nursery & 0.560 & 12958 & 9 & human-behaviour \\
gametes\_epistasis & 0.521 & 1600 & 21 & deterministic-simulated \\
mofn-3-7-10 & 0.504 & 1324 & 11 & deterministic-simulated \\
parity5\_plus\_5 & 0.444 & 1124 & 11 & deterministic-simulated \\
\end{longtable}
\endgroup

\begingroup
\setlength{\tabcolsep}{1pt}
\begin{longtable}{
@{}
>{\raggedright\arraybackslash}p{0.34\textwidth}
r
r
r
>{\raggedright\arraybackslash}p{0.24\textwidth}
@{}
}
\caption{$R^2$ scores on CTR-23 for TabDPT pre-trained on various datasets, sorted in descending order of $R^2$.}\label{tab:ctr23_r2}\\
\toprule
\textbf{Dataset} & \textbf{$R^2$ on CTR-23} & \textbf{\# Instances} & \textbf{\# Features} & \textbf{Dataset Domain} \\
\midrule
\endfirsthead
\caption[]{$R^2$ scores on CTR-23 for TabDPT pre-trained on various datasets, sorted in descending order of $R^2$.}\\
\toprule
\textbf{Dataset} & \textbf{$R^2$ on CTR-23} & \textbf{\# Instances} & \textbf{\# Features} & \textbf{Dataset Domain} \\
\midrule
\endhead

\bottomrule
\endfoot
colleges & 0.693 & 7063 & 45 & other \\
kick & 0.693 & 72983 & 33 & industrial-operational \\
ipums\_la\_97-small & 0.691 & 7019 & 61 & financial-demographic \\
dionis & 0.685 & 416188 & 61 & other \\
ipums\_la\_98-small & 0.685 & 7485 & 56 & financial-demographic \\
Census-Income & 0.684 & 299285 & 42 & financial-demographic \\
APSFailure & 0.682 & 76000 & 171 & industrial-operational \\
ipums\_la\_99-small & 0.681 & 8844 & 57 & financial-demographic \\
KDDCup09\_churn & 0.679 & 50000 & 231 & industrial-operational \\
covertype & 0.679 & 581012 & 55 & biology-ecology \\
road-safety & 0.678 & 111762 & 33 & human-behaviour \\
gas-drift-different-concentrations & 0.677 & 13910 & 130 & other-science \\
gas-drift & 0.672 & 13910 & 129 & other-science \\
eye\_movements & 0.672 & 10936 & 28 & medical-human-sensor \\
musk & 0.671 & 6598 & 168 & other-science \\
KDDCup09\_appetency & 0.671 & 50000 & 231 & human-behaviour \\
volkert & 0.669 & 58310 & 181 & other \\
MiniBooNE & 0.668 & 130064 & 51 & physics-astronomy \\
sylvine & 0.667 & 5124 & 21 & other \\
cardiotocography & 0.665 & 2126 & 36 & medical-human-sensor \\
dilbert & 0.665 & 10000 & 2001 & other \\
default-of-credit-card-clients & 0.664 & 13272 & 21 & financial-demographic \\
sylva\_agnostic & 0.657 & 14395 & 217 & biology-ecology \\
helena & 0.655 & 65196 & 28 & other \\
jannis & 0.649 & 83733 & 55 & other \\
pol & 0.647 & 10082 & 27 & industrial-operational \\
scene & 0.647 & 2407 & 300 & vision-audio-text \\
philippine & 0.646 & 5832 & 309 & other \\
heloc & 0.645 & 10000 & 23 & financial-demographic \\
one-hundred-plants-texture & 0.643 & 1599 & 65 & biology-ecology \\
magic & 0.643 & 19020 & 11 & physics-astronomy \\
elevators & 0.641 & 16599 & 19 & other \\
MagicTelescope & 0.640 & 19020 & 12 & physics-astronomy \\
house\_16H & 0.639 & 22784 & 17 & financial-demographic \\
PieChart3 & 0.638 & 1077 & 38 & other \\
ldpa & 0.635 & 164860 & 8 & medical-human-sensor \\
JapaneseVowels & 0.634 & 9961 & 15 & vision-audio-text \\
page-blocks & 0.633 & 5473 & 11 & vision-audio-text \\
christine & 0.633 & 5418 & 1637 & other \\
spoken-arabic-digit & 0.631 & 263256 & 15 & vision-audio-text \\
higgs & 0.630 & 98050 & 29 & physics-astronomy \\
eeg-eye-state & 0.630 & 14980 & 15 & medical-human-sensor \\
colleges\_usnews & 0.629 & 1302 & 34 & other \\
jasmine & 0.627 & 2984 & 145 & other \\
SpeedDating & 0.627 & 8378 & 121 & human-behaviour \\
Satellite & 0.623 & 5100 & 37 & physics-astronomy \\
cjs & 0.621 & 2796 & 35 & biology-ecology \\
porto-seguro & 0.620 & 595212 & 58 & human-behaviour \\
PizzaCutter3 & 0.617 & 1043 & 38 & other \\
Click\_prediction\_small & 0.615 & 39948 & 12 & human-behaviour \\
ada\_prior & 0.612 & 4562 & 15 & financial-demographic \\
shuttle & 0.608 & 58000 & 10 & physics-astronomy \\
sf-police-incidents & 0.607 & 2215023 & 9 & human-behaviour \\
artificial-characters & 0.604 & 10218 & 8 & deterministic-simulated \\
pbcseq & 0.585 & 1945 & 19 & medical-human-sensor \\
airlines & 0.585 & 539383 & 8 & industrial-operational \\
house\_8L & 0.581 & 22784 & 9 & financial-demographic \\
ada & 0.581 & 4147 & 49 & other \\
kdd\_internet\_usage & 0.579 & 10108 & 69 & financial-demographic \\
kropt & 0.578 & 28056 & 7 & deterministic-simulated \\
analcatdata\_halloffame & 0.575 & 1340 & 17 & other \\
albert & 0.575 & 425240 & 79 & other \\
chess & 0.574 & 28056 & 7 & deterministic-simulated \\
ada\_agnostic & 0.571 & 4562 & 49 & financial-demographic \\
okcupid-stem & 0.567 & 50789 & 20 & human-behaviour \\
guillermo & 0.565 & 20000 & 4297 & other \\
kr-vs-k & 0.560 & 28056 & 7 & deterministic-simulated \\
walking-activity & 0.558 & 149332 & 5 & medical-human-sensor \\
credit & 0.558 & 16714 & 11 & financial-demographic \\
compas-two-years & 0.557 & 4966 & 12 & human-behaviour \\
visualizing\_soil & 0.545 & 8641 & 5 & biology-ecology \\
fabert & 0.544 & 8237 & 801 & other \\
analcatdata\_supreme & 0.543 & 4052 & 8 & other \\
poker-hand & 0.512 & 1025009 & 11 & deterministic-simulated \\
Diabetes130US & 0.500 & 71090 & 8 & medical-human-sensor \\
fbis.wc & 0.483 & 2463 & 2001 & vision-audio-text \\
madeline & 0.478 & 3140 & 260 & other \\
pollen & 0.424 & 3848 & 6 & biology-ecology \\
puma8NH & 0.367 & 8192 & 9 & deterministic-simulated \\
twonorm & 0.316 & 7400 & 21 & deterministic-simulated \\
yeast & 0.278 & 1269 & 9 & biology-ecology \\
hill-valley & 0.221 & 1212 & 101 & deterministic-simulated \\
mofn-3-7-10 & 0.131 & 1324 & 11 & deterministic-simulated \\
mushroom & 0.112 & 8124 & 23 & biology-ecology \\
parity5\_plus\_5 & -0.048 & 1124 & 11 & deterministic-simulated \\
nursery & -0.228 & 12958 & 9 & human-behaviour \\
gametes\_epistasis & -0.806 & 1600 & 21 & deterministic-simulated \\
imdb.drama & -16.712 & 120919 & 1002 & vision-audio-text \\
\end{longtable}
\endgroup

\section{Training and architecture details}
\label{app:training}

Throughout this paper, ``TabDPT'' refers specifically to TabDPT v1.1.

\subsection{Base model: many-dataset pre-training ablations}
\label{app:training:base}

This section provides the experimental setup for the many-dataset pre-training ablations. The deduplication, dataset-level filtering, and column-level pre-processing experiments all use the same base TabDPT architecture and training configuration described henceforth.

\paragraph{Pre-training corpus} We use part of the datasets collected by~\citet{hosseinzadeh2026tabdpt} as our pre-training corpus. This contains $1{,}732$ datasets collected from OpenML~\cite{OpenML2025}, with datasets contained in the CC-18, CTR-23, or TabArena~\cite{erickson2025tabarena} benchmarks automatically removed. We do not filter datasets with fewer than 10 columns, unlike the cited paper, but we do exclude datasets with no clear target variable $y$. We additionally find that $197$ of these datasets are exact duplicates. We cut off each dataset at $10{,}000$ rows and $3{,}000$ columns to bound pre-processing costs.

\paragraph{Architecture} We use a 12-layer transformer with embedding dimension $d_{\text{model}}=512$, 8 attention heads, and MLP hidden dimension 1{,}024 (hidden factor 2). The model uses post-layer normalization and an asymmetric in-context attention pattern: query (test) tokens attend only to context (train) tokens, never to each other. Feature columns are projected to $d_{\text{model}}$ via a shared linear encoder. For context tokens, the target embedding (a separate learned linear layer) is added on top of the feature embedding. At each step, inputs are capped at 100 feature columns and 10 output classes.

\paragraph{Training} At each step, one column is selected uniformly at random as the prediction target $y$ while the remaining columns form the feature matrix $X$. Classification and regression heads are trained jointly in a single forward pass using a self-supervised SSL objective. Label smoothing of $0.1$ is applied to the classification cross-entropy loss. We use AdamW with learning rate $5\times10^{-4}$, weight decay $0.05$, and no learning-rate schedule. Gradient norms are clipped to $1.0$. Dropout is $0.0$ throughout. Each run trains for 128{,}000 gradient update steps, with per-step batch size of 32 and gradient accumulation over 8 steps for an effective batch size of 256. The context window is randomly sampled at each step and is at minimum 50 tokens and at maximum 1{,}024.

\paragraph{Data pre-processing and retrieval} Each dataset is truncated to at most 10{,}000 rows and 3{,}000 columns before any pre-processing. Categorical columns are coerced to integers via label encoding and columns where fewer than 1\% of values are interpretable as numeric are treated as categorical. Missing values are imputed with column means. The joint $[X \mid y]$ matrix is then standardized with a \texttt{StandardScaler}. A FAISS flat L2 index is built on training-split embeddings and used for retrieval-augmented context assembly during training.

\paragraph{Evaluation} We report results on the same two standard benchmarks used previously: \textbf{CC-18} for classification and \textbf{CTR-23} for regression. On all metrics, we report the \textbf{Interquartile Mean} (IQM), which trims the bottom and top quartiles and is robust to outliers. We also report $95\%$ bootstrap confidence intervals computed via stratified bootstrap over datasets using $20{,}000$ resampling iterations. Inference uses context size $2{,}048$ and $8$ ensemble passes on the raw benchmark datasets, without applying any column pre-processing, to ensure consistency across experiments.

\paragraph{Hardware} Each run uses a single Nvidia A100 (40\,GB) GPU with 8 CPU workers and 150GB RAM.

\subsection{Large model: context-length and architecture scaling}
\label{app:training:large}

The experiments in subsection~\ref{sec:larger} use a larger model trained at three
context lengths, with and without pre-processing, for six runs total. All six share the configuration described here.

\paragraph{Architecture} We increase the model to 16 transformer layers and embedding dimension $d_{\text{model}}=768$, with 8 attention heads and MLP hidden dimension 1{,}536 (hidden factor 2). Pre-layer normalization replaces the post-norm used in the base model. All other architectural choices (asymmetric attention, feature/label encoders,
output heads, input caps) are unchanged. 

\paragraph{Training} We use the ScheduleFree AdamW optimizer with learning rate $1\times10^{-3}$ and weight decay $0.05$. Training uses BF16 mixed precision. Gradient clipping ($1.0$) and label smoothing ($0.1$) are unchanged. Each run trains for 128{,}000 gradient steps. Batch size is 32 with gradient accumulation over 4 steps, giving effective batch size 128. The three variants differ in the maximum number of labelled context tokens (shown in the main text) but the minimum context size is 100 for all three. Inference for all six runs uses context size $2{,}048$ with no column pre-processing applied, for consistency across comparisons.

\paragraph{Data pre-processing and retrieval} Dataset caps, categorical encoding, imputation, scaling, and FAISS L2 retrieval are identical to the base model.
For the ``$+$pp'' variants, the NaN-column drop, Spearman correlation deduplication
(threshold $0.90$), and minimum-feature filter ($k=5$) are applied to each dataset
before training, with the column mask fitted on the training split as in the base
ablations.

\paragraph{Hardware} Each of the six runs uses a single Nvidia H100 GPU. All other software and infrastructure settings are as described in
Section~\ref{app:training:base}.

\section{Additional TabArena results}
\label{app:tabarena}

We provide extended results on the TabArena benchmark~\citep{erickson2025tabarena}. All experiments use TabArena-Lite (51 tasks, fold 0), bootstrapped 2000 times. Elo is computed against a pool of default-configuration baselines (no tuning or ensembling), and is anchored at Random Forest = 1000. We evaluate six TabDPT variants spanning a $2 \times 3$ grid over pre-processing ($+$pp vs.\ none) and training context length (1k, 2k, 4k tokens), alongside a broad set of baseline methods.

\subsection{Pairwise win rates}

\begin{figure}[t]
    \centering
    \includegraphics[width=0.9\linewidth]{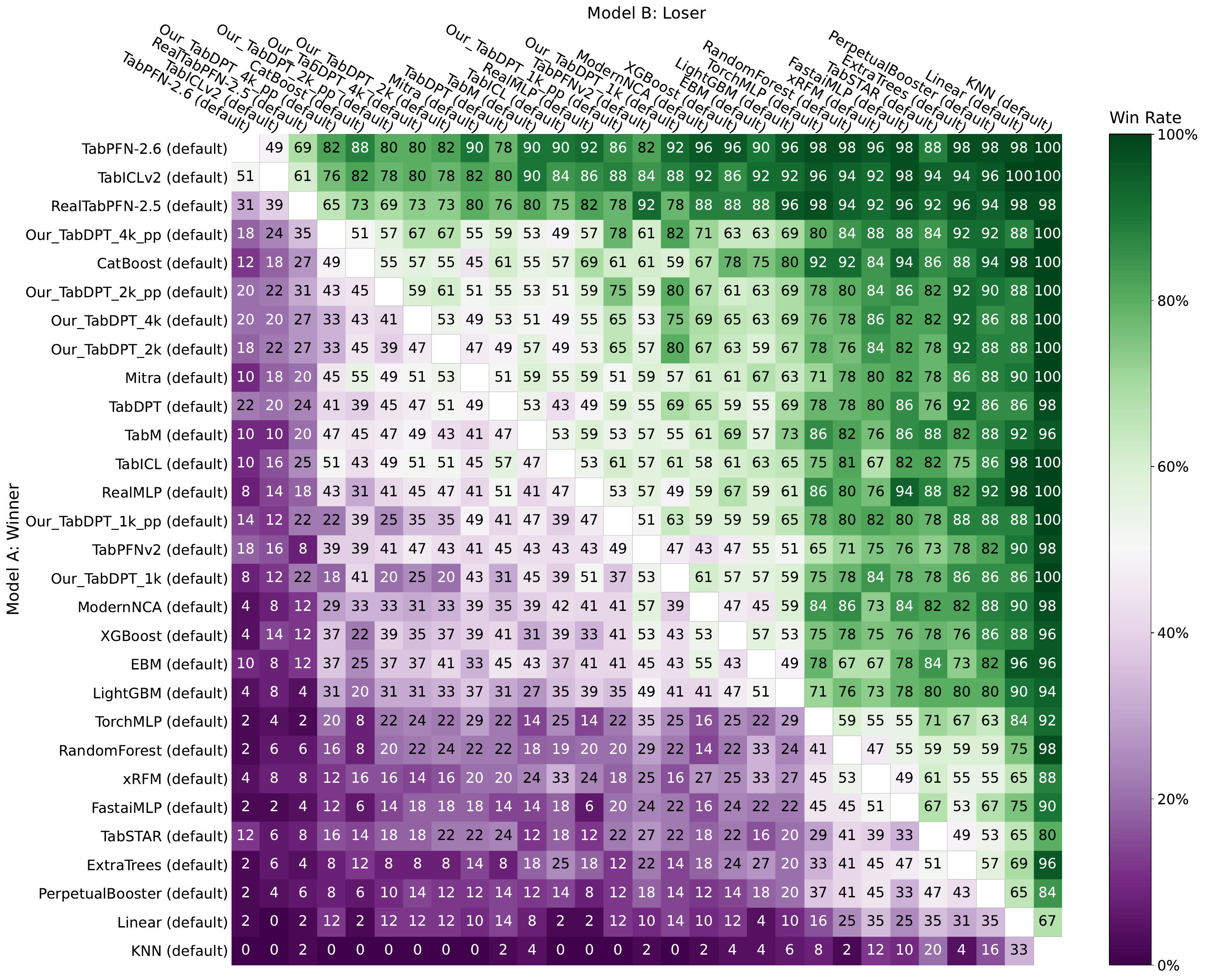}
    \caption{Pairwise win-rate matrix on TabArena-Lite across all evaluated methods (our models and baselines). Each cell shows the fraction of tasks where the row method outperforms or ties the column method.}
    \label{fig:app_pairwise_winrate}
\end{figure}

Figure~\ref{fig:app_pairwise_winrate} shows pairwise win rates across all methods. The relative ordering among our variants matches the main text: 4k$+$pp performs best, followed by 2k$+$pp and 4k, with shorter-context models trailing.

\subsection{Leaderboard results}

\begin{table}[t]
\centering
\caption{Full TabArena-Lite leaderboard (51 tasks, fold 0). Elo is anchored at Random Forest = 1000. Our models are in bold.}
\label{tab:app_tabarena}
\begin{tabular}{lcc}
\toprule
Method & Elo & Avg. Rank \\
\midrule
TABPFN-V2.6 & 1583 & 4.16 \\
TABICLv2 & 1545 & 4.80 \\
REALTABPFN-V2.5 & 1477.6 & 6.16 \\
\textbf{Our TabDPT (4k$+$pp)} & \textbf{1322.8} & \textbf{10.25} \\
CatBoost & 1320.8 & 10.31 \\
\textbf{Our TabDPT (2k$+$pp)} & \textbf{1299.8} & \textbf{10.96} \\
\textbf{Our TabDPT (4k)} & \textbf{1274.3} & \textbf{11.76} \\
\textbf{Our TabDPT (2k)} & \textbf{1270.7} & \textbf{11.88} \\
MITRA\_GPU & 1265.8 & 12.04 \\
TabDPT\_GPU & 1259.1 & 12.25 \\
TabM\_GPU & 1258.4 & 12.27 \\
TabICL\_GPU & 1257.2 & 12.31 \\
RealMLP\_GPU & 1245.1 & 12.71 \\
\textbf{Our TabDPT (1k$+$pp)} & \textbf{1220.0} & \textbf{13.53} \\
TabPFNv2\_GPU & 1195.6 & 14.33 \\
\textbf{Our TabDPT (1k)} & \textbf{1190.8} & \textbf{14.49} \\
ModernNCA\_GPU & 1187.0 & 14.62 \\
XGBoost & 1179.5 & 14.86 \\
EBM & 1177.1 & 14.94 \\
LightGBM & 1151.5 & 15.78 \\
NeuralNetTorch & 1024.4 & 19.75 \\
Random Forest (anchor) & 1000.0 & 20.44 \\
xRFM\_GPU & 999.6 & 20.45 \\
FastAI & 973.1 & 21.18 \\
TabSTAR & 954.4 & 21.67 \\
ExtraTrees & 945.9 & 21.88 \\
PerpetualBooster & 908.8 & 22.78 \\
Linear Model & 816.1 & 24.71 \\
KNN & 583.2 & 27.71 \\
\bottomrule
\end{tabular}
\end{table}

Table~\ref{tab:app_tabarena} reports leaderboard metrics for all evaluated methods. The 4k$+$pp variant achieves the strongest performance among our models, and the remaining variants follow the same ordering as in Figure~\ref{fig:app_pairwise_winrate}.

\section{Attention similarity experiments: methods and supplementary results}
\label{app:attn}

\subsection{Models, probes, and similarity calculations}

\paragraph{Checkpoints} We probe 52 TabDPT checkpoints, each pre-trained from scratch on a single OpenML dataset, all sharing the same architecture (12 transformer blocks, $d_{\text{model}}=512$, single attention head, asymmetric in-context attention pattern) and the same training schedule (lr $5\times 10^{-4}$, weight decay $0.05$, sequence length 1536, taken at 12{,}800 gradient updates with an effective batch size of 256).

\paragraph{Probes} Each model is evaluated on two held-out OpenML classification datasets, \textsc{Amazon\_employee\_access} (OpenML ID 4135) and \textsc{nursery} (OpenML ID 1568), that none of the 52 models was trained on. For each (model, probe) we record the per-block attention tensor of shape $(B,H,Q,C)=(32,1,512,512)$ at a fixed in-context cutoff $\mathrm{eval\_pos}=512$, where $\alpha_{b,h,q,c}=\mathrm{softmax}_c(\langle\mathbf{q}_{b,h,q}, \mathbf{k}_{b,h,c}\rangle/\sqrt{d_{\text{head}}})$ is a probability vector over the $C$ context tokens (the \emph{attention distribution} of query $q$).

\paragraph{Pair similarity} For two models $i,j$ on probe $p$ and block $\ell$ we compute
\begin{equation*}
  s_{ij}^{(\ell,p)} = \tfrac{1}{B Q}\sum_{b,q}
  \rho_{\text{Spearman}}\!\big(\alpha^{(i)}_{b,q,\cdot},\,
  \alpha^{(j)}_{b,q,\cdot}\big),
\end{equation*}
which is the row-wise Spearman rank correlation between the two attention distributions, that is then averaged over queries and probe examples. The pair-level scalar in the main paper averages $s_{ij}^{(\ell,p)}$ over the 12 blocks. Spearman is preferred over Pearson because rank correlation is invariant to the exact magnitudes of the attention values.

\paragraph{Held-out AUC} Per-model AUC is the AUC on the OpenML CC-18 classification suite at the same training step, which is data disjoint from both the training datasets and the two probes.

\subsection{Strong / weak partitioning}

We split the 52 models by AUC quartile ($q_{25}=0.831$, $q_{75}=0.900$). Models with $\text{AUC}\ge q_{75}$ are \textbf{strong} (S, 13 models), models with $\text{AUC}\le q_{25}$ are \textbf{weak} (W, 13 models). The $26$ middle-quartile models are excluded.

\begin{table}[ht]
  \centering
  \caption{Models in the strong and weak sets, identified by their training dataset.}
  \label{tab:attn-quartile-membership}
  \begin{tabular}{ll}
    \toprule
    \textbf{Strong (AUC $\ge 0.900$)} &
    \textbf{Weak (AUC $\le 0.831$)} \\
    \midrule
    \textsc{APSFailure}, \textsc{Census-Income}, &
      \textsc{Diabetes130US}, \textsc{airlines}, \\
    \textsc{KDDCup09\_appetency}, \textsc{SpeedDating}, &
      \textsc{analcatdata\_supreme}, \textsc{chess}, \\
    \textsc{cardiotocography}, \textsc{Colleges}, &
      \textsc{hill-valley}, \textsc{parity5\_plus\_5}, \\
    \textsc{covertype}, \textsc{gas-drift}, &
      \textsc{pbcseq}, \textsc{poker-hand}, \\
    \textsc{ipums\_la\_98-small}, \textsc{musk}, &
      \textsc{pollen}, \textsc{puma8NH}, \\
    \textsc{pol}, \textsc{road-safety}, \textsc{scene} &
      \textsc{twonorm}, \textsc{visualizing\_soil}, \textsc{yeast} \\
    \bottomrule
  \end{tabular}
\end{table}

All $95\%$ CIs in Tables~\ref{tab:attn-summary} and~\ref{tab:attn-perblock-appendix} are obtained by bootstrap. We repeatedly draw a version of each group by sampling its observations with replacement, recompute the gap between the two group means, and collect the resulting gap estimates. The $95\%$ CI is the range from the $2.5$th to the $97.5$th percentile of that collection. We use $10\,000$ resamples for Table~\ref{tab:attn-summary} and $2\,000$ for Table~\ref{tab:attn-perblock-appendix}. The random seed $0$ is fixed for reproducibility.

\subsection{Per-block breakdown}

Table~\ref{tab:attn-perblock-appendix} reports, per transformer block, the SS / SW / WW group means together with the SS$-$SW gap and its $95\%$ bootstrap CI ($K=2\,000$). Each row pools across the two probes, so $n_{\text{SS}}=156$, $n_{\text{SW}}=338$, $n_{\text{WW}}=156$ per block. 

\begin{table}[h]
  \centering
  \caption{Per-block group means and SS$-$SW gap with $95\%$ bootstrap CI. One observation per (model pair, probe, layer). Rows whose CI excludes zero are marked $\dagger$.}
  \label{tab:attn-perblock-appendix}
  \begin{tabular}{cccccc}
    \toprule
    Block & SS & SW & WW & SS$-$SW & $95\%$ CI \\
    \midrule
    $0$  & $0.854$ & $0.696$ & $0.593$ & $+0.158^{\dagger}$ & $[+0.124,+0.193]$ \\
    $1$  & $0.525$ & $0.442$ & $0.402$ & $+0.083^{\dagger}$ & $[+0.005,+0.164]$ \\
    $2$  & $0.123$ & $0.047$ & $0.279$ & $+0.075$           & $[-0.027,+0.176]$ \\
    $3$  & $0.145$ & $0.110$ & $0.156$ & $+0.035$           & $[-0.050,+0.119]$ \\
    $4$  & $0.551$ & $0.383$ & $0.330$ & $+0.168^{\dagger}$ & $[+0.132,+0.205]$ \\
    $5$  & $0.584$ & $0.363$ & $0.238$ & $+0.221^{\dagger}$ & $[+0.187,+0.255]$ \\
    $6$  & $0.577$ & $0.403$ & $0.277$ & $+0.174^{\dagger}$ & $[+0.133,+0.218]$ \\
    $7$  & $0.303$ & $0.295$ & $0.325$ & $+0.008$           & $[-0.057,+0.075]$ \\
    $8$  & $0.520$ & $0.368$ & $0.269$ & $+0.153^{\dagger}$ & $[+0.124,+0.181]$ \\
    $9$  & $0.585$ & $0.363$ & $0.253$ & $+0.222^{\dagger}$ & $[+0.197,+0.249]$ \\
    \textbf{$10$} & ${0.617}$ & ${0.213}$ & ${0.073}$
                 & ${+0.404^{\dagger}}$ & ${[+0.367,+0.439]}$ \\
    $11$ & $0.667$ & $0.407$ & $0.288$ & $+0.260^{\dagger}$ & $[+0.239,+0.284]$ \\
    \bottomrule
  \end{tabular}
\end{table}

The SS$-$SW gap is positive in all 12 blocks, statistically significant in 9 of them (blocks $0$, $1$, $4$, $5$, $6$, $8$, $9$, $10$, $11$), and largest at block $10$.

\begin{figure}[t]
\centering
\includegraphics[width=0.7\linewidth]{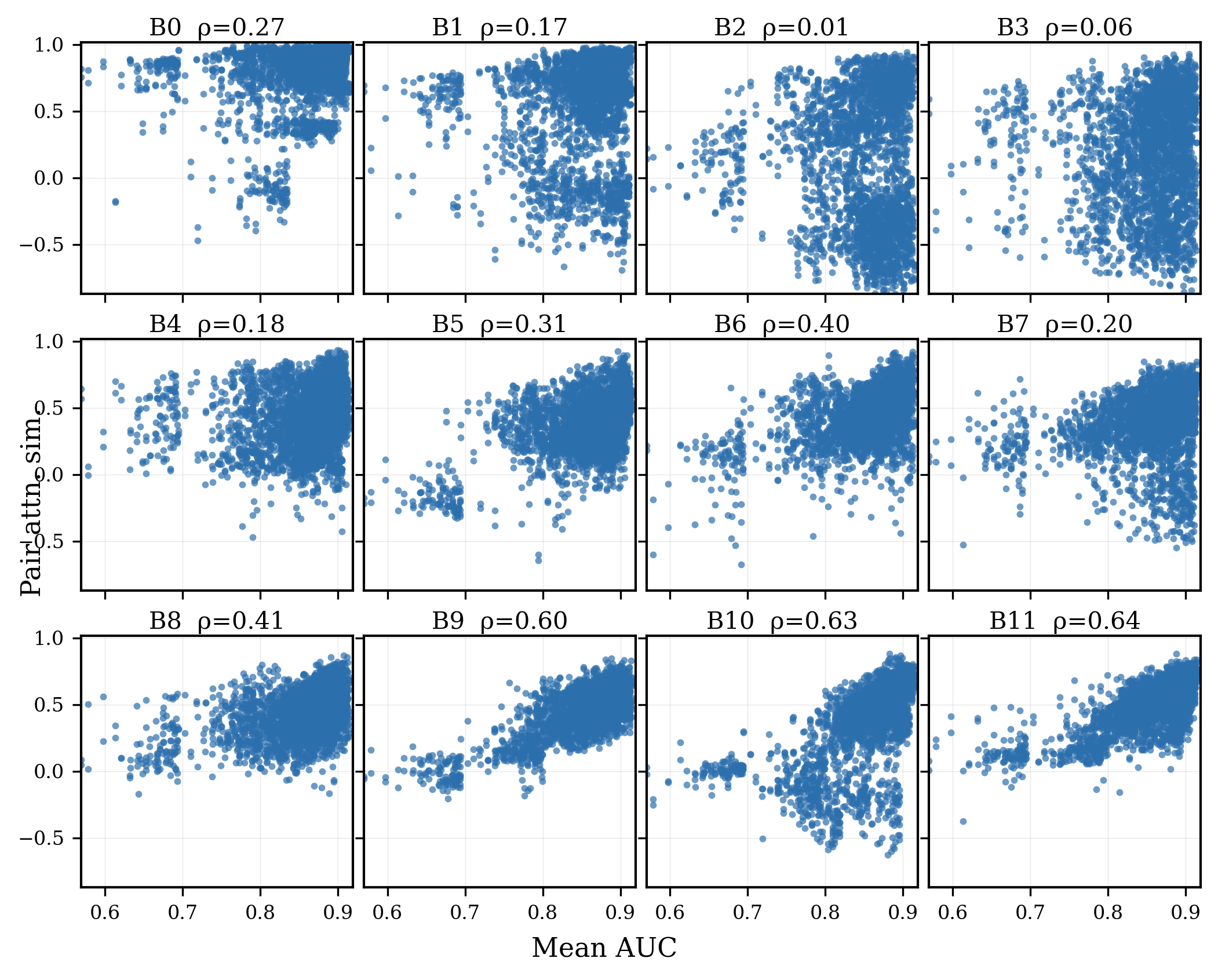}
\caption{Per-block pair attention similarity vs.\ mean AUC, $n=2652$ points (one per model pair $\times$ probe dataset). %
\label{fig:pair-auc-attn-bylayer}
}
\end{figure}

We also see the global correlation effect shown in Figure~\ref{fig:pair-auc-attn-overall}, being concentrated in the deeper transformer blocks in Figure~\ref{fig:pair-auc-attn-bylayer}: per-block correlation rises from near zero in early blocks to $\approx 0.6$ in blocks $9$--$11$.

\subsection{Caveat: pair non-independence}

The data points are not fully independent, since each model appears in $51$ pairs, so the confidence intervals reported above are somewhat tighter than they would be under a stricter resampling scheme.

\section{Additional query--key sharing ablation}
\label{app:qk_extra}

\begin{figure}[t]
\centering
\begin{minipage}{.44\textwidth}
    \centering
    \includegraphics[width=0.85\linewidth]{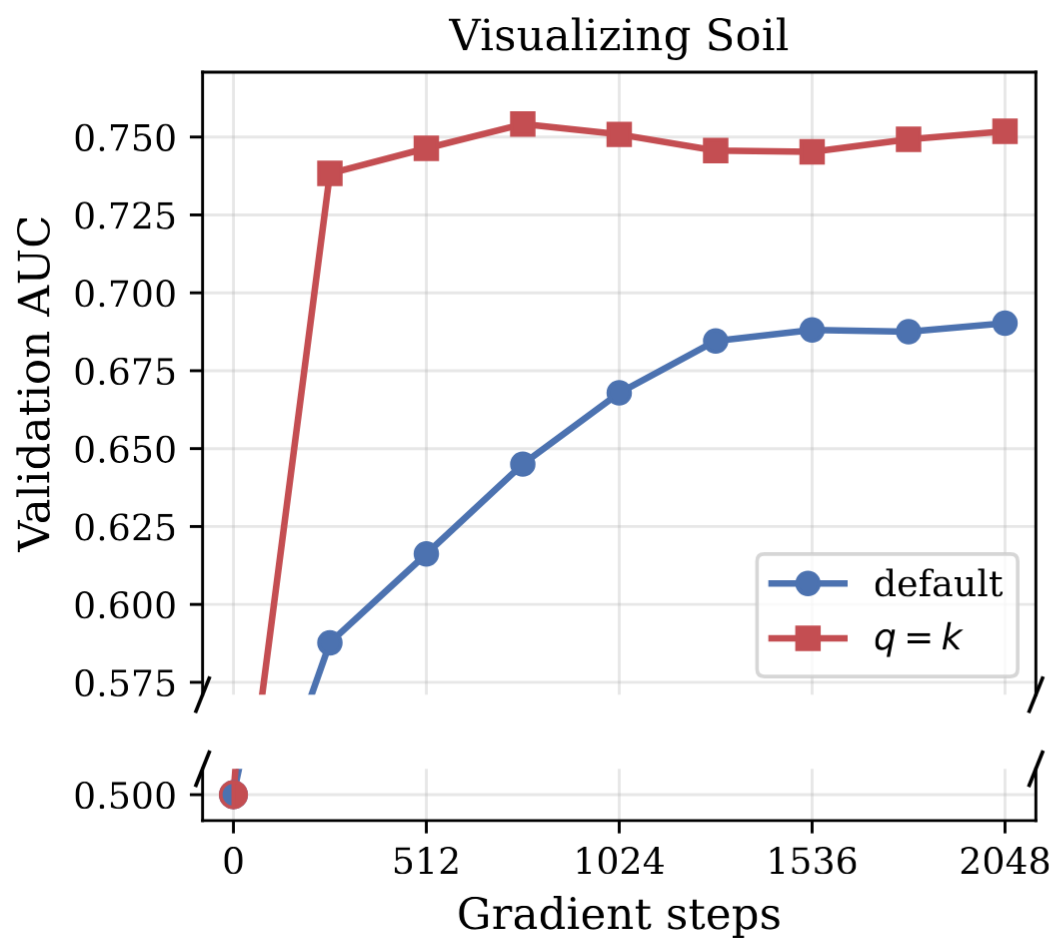}
\end{minipage}
\hspace{0.04\textwidth}
\begin{minipage}{.44\textwidth}
    \centering
    \includegraphics[width=0.85\linewidth]{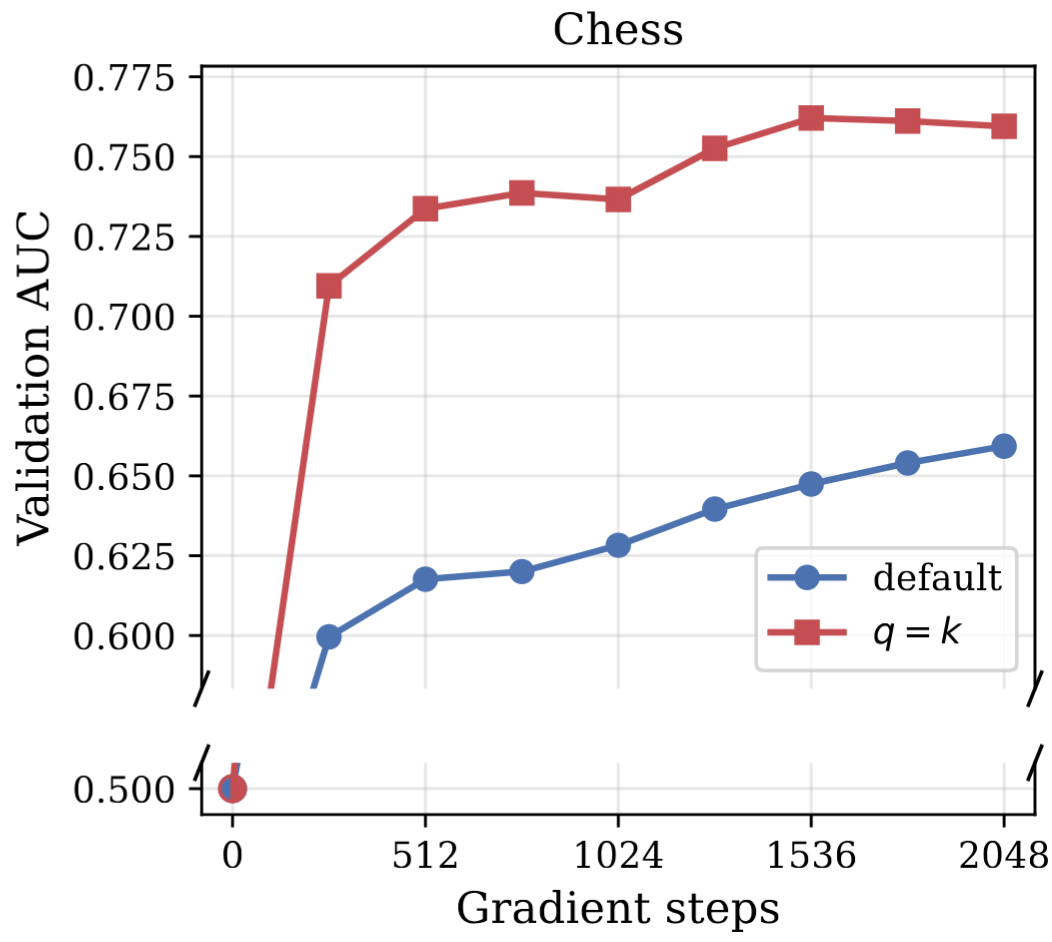}
\end{minipage}
\caption{\looseness=-1
    Ablations where $W_Q = W_K$ is enforced, on \textsc{Visualizing Soil} (left) and \textsc{Chess} (right), using the same setting as Figure~\ref{fig:qk_ablation_weak}.
    }
\label{fig:qk_ablation_extra}
\end{figure}

Figure~\ref{fig:qk_ablation_extra} is an additional analysis of training dynamics when $W_Q=W_K$ is enforced, on two further pre-training tables that generalize poorly, the \textsc{Visualizing Soil} and the \textsc{Chess} datasets. 
In both cases, forcing $W_Q=W_K$ yields significantly higher validation AUC than the default architecture, complementing the main-text weak-table results of Figure~\ref{fig:qk_ablation_weak} and supporting the view that TFM performance is close to a nearest neighbour-like mechanism.

\clearpage

\end{document}